%% file: main_arxiv.tex
\documentclass{uai2025_arxiv} 

\usepackage[american]{babel}

\usepackage{natbib} 
\usepackage{mathtools} 
\usepackage{booktabs} 
\usepackage{tikz} 

\usepackage{dsfont, bm, amsmath, amssymb}
\usepackage{graphicx, subcaption, booktabs, multirow}
\usepackage{mathtools}
\mathtoolsset{showonlyrefs}
\newcommand{\blind}{1} 

\title{Targeted synthetic data generation for tabular data via hardness characterization}

\author[1]{Tommaso Ferracci}
\author[2]{Leonie Tabea Goldmann}
\author[2]{Anton Hinel}
\author[1]{\href{mailto:<f.sannapassino@imperial.ac.uk>}{Francesco Sanna Passino}{}}
\affil[1]{%
    Department of Mathematics, Imperial College London
}
\affil[2]{%
Decision Science, Credit \& Fraud Risk, American Express
}
  
\begin{document}
\maketitle

\begin{abstract}
\input{sections/abstract}
\end{abstract}

\input{sections/intro}

\input{sections/background}

\input{sections/methods}
\input{sections/simulations}

\input{sections/results}

\input{sections/discussion}


\input{sections/reproducibility}

\bibliography{refs}

\newpage
\appendix
\onecolumn

\title{APPENDIX \\ Targeted synthetic data generation for tabular data via hardness characterization}

\maketitle

\section{Training details}
\input{sections/appendix}

\end{document}

%% file: sections/abstract.tex
Data augmentation via synthetic data generation has been shown to be effective in improving model performance and robustness in the context of scarce or low-quality data. 
Using the data valuation framework to statistically identify beneficial and detrimental observations, we introduce a simple 
augmentation pipeline that generates only high-value training points based on hardness characterization, in a computationally efficient manner.
We first empirically demonstrate via benchmarks on real data that Shapley-based data valuation methods perform comparably with learning-based methods in hardness characterization tasks, while offering significant 
computational advantages. Then, we show that synthetic data generators trained on the hardest points outperform non-targeted data augmentation on a number of tabular datasets. Our approach improves the quality of out-of-sample predictions and it is  computationally more efficient compared to non-targeted methods.

%% file: sections/intro.tex
\section{Introduction}\label{sec:intro}

Training complex machine learning models requires large amounts of data, but in real-world applications, data may be of poor quality, insufficient in amount, or subject to privacy, safety, and regulatory limitations. Such challenges have sparked an interest in \textit{synthetic data generators} (SDGs), which use available training data to generate realistic synthetic samples \citep{review}.
In this work, we argue that when the objective is to use synthetic data to make an existing machine learning model better generalize to unseen data, augmenting only the hardest training points is more effective and has significant computational advantages compared to augmenting the entire training dataset. In other words, we propose a novel \textit{scalable targeted synthetic data generation} framework, focusing on binary classification tasks. The underlying intuition is that within a dataset, some of the observations are obvious to classify, whereas others play a more significant role in determining the decision boundary of the trained model. Focusing only on these harder examples when generating additional data may allow the model to efficiently learn a more robust decision boundary with the additional advantage of reducing computational costs associated with the training of a synthetic data generator. 

In particular, the method proposed in this work is specifically targeted toward binary classifiers on tabular data, which often require adaptations such as oversampling, undersampling, or synthetic data generation to improve model performance under class imbalance \citep{smote, He08}. Although deep neural networks have proven successful across domains such as images, audio, or text, they are still regularly outperformed on tabular data by simpler and more interpretable tree-based architectures \citep{shwartz-ziv_tabular_2021}. Therefore, we choose trees as the reference model for binary classification on tabular data, while deep learning models are used for synthetic data generation. In general, we will use $D = \{(\bm{x}_i, y_i)\}_{i=1}^n$ to denote the training data, where $\bm{x}_i \in \mathbb{R}^d$ is a $d$-dimensional feature vector and $y_i \in \{0, 1\}$ is a binary label.

Our targeted synthetic data generation approach is divided into two main steps: \textit{(i)} hardness characterization of training data points, followed by \textit{(ii)} training of a synthetic data generator only on the hardest data points of the initial training dataset. In order to identify ``hard'' data points, we propose a novel hardness characterizer specific to tabular data, based on KNN data Shapleys \citep{knn_shapley}. We show that KNN data Shapleys are capable of achieving performances comparable to state-of-the-art methods in hardness detection benchmarks, while offering key advantages such as being deterministic, model agnostic and, most importantly, computationally efficient when compared to existing alternatives in the literature. 
The computational advantages of the proposed data augmentation pipeline mainly stem from two aspects: \textit{(i)} the identification of ``hard'' data points via KNN Shapleys, a computationally efficient method against other hardness characterizers; and \textit{(ii)} the reduction of the number of data points used for training of synthetic data generators, focusing only on such ``hard'' data points. 
Even though hardness characterization has already been applied in the literature with rationales similar to this work, the focus was mostly on pruning the easier examples \citep{scalability, grand}, rather than augmenting the harder ones. 

Once the hardest points have been identified, synthetic data generation models are trained only on the most difficult points identified by our hardness characterizer. In this work, we focus particularly on two widely used deep 
architectures for synthetic data generation for tabular data, namely Tabular Variational Autoencoders (TVAE) and Conditional Tabular Generative Adversarial Networks (CTGAN) \citep[both introduced in][]{ctgan}, but we stress that our pipeline can be applied to \textit{any} synthetic data generator. 
When comparing the best models on various different datasets given a fixed number of points to generate, we find that augmenting only hard data points results in a larger performance improvement on unseen data with respect to non-targeted data augmentation, while being computationally cheaper.

The rest of the work is structured as follows: after summarizing the related literature in Section \ref{sec:background}, the proposed approach is discussed and empirically verified in Section~\ref{sec:methods}. 
Next, we explore the use of KNN Shapleys as hardness characterizers via a benchmark study in Section~\ref{sec:benchmark}, followed by an outline on how to utilize them for synthetic data generation on real (Sections~\ref{sec:amex_res} and~\ref{sec:add_res}) and simulated data (Section~\ref{sec:sim_ex}). The results are discussed in Section~\ref{sec:discussion}.

%% file: sections/background.tex
\section{Background and related literature} \label{sec:background}

In this work, we aim to bridge the gap between existing methods for hardness characterization and game theoretic approaches to data valuation, with the objective of improving the performance of synthetic data generators. Therefore, our work aligns with existing techniques in hardness characterization, data valuation, and synthetic data generation. 

\subsection{Hardness characterization} \label{sec:hard_cat}

As described in the introduction, the first aim of this work is the identification of a subset of points that are hard for the model to classify. Existing literature on hardness characterization is often qualitative and based on different definitions of ``hardness''. Following the taxonomy presented in \cite{hard_cat}, three main categories of hardness can be identified: \textit{(i) Mislabeling}: the true label is replaced with an incorrect one. The label perturbation probability can be uniform across classes 
    or label-dependent;
\textit{(ii) Out-of-distribution}: features are transformed or shifted, leading to observations distributed differently from the main data generating process;
\textit{(iii) Atypical}: observations that are compatible with the main data generating process, but located in tails or rare regions of the feature distributions.
These distinctions normally dictate different courses of actions: datapoints that are mislabeled or out of distribution are symptomatic of errors in the data collection process and should be removed from training data, whereas atypical data 
may 
need to be augmented to train a robust model \citep{Yuksekgonul23}.

While many different hardness characterizers have been proposed, most of them are based around keeping track of some statistic of interest for each datapoint during the training process, meaning they can be adapted to any model that is trained iteratively (such as XGBoost or a neural network). Examples include \textit{GraNd} \citep{grand}, which uses large gradients as an indicator of hardness, and \textit{Forget} \citep{forget}, which counts how many times each point is learned and then forgotten during training, with the simple intuition that hard points are forgotten more often. Based on the quantitative benchmarks reported in \cite{hard_cat}, \textit{Data-IQ} \citep{dataiq} and CleanLab \citep{Northcutt21} are often the best performers on tabular data, and they 
can be considered as benchmark hardness characterizers for tabular data 
because of their interpretability and compatibility with XGBoost.



\subsection{Data valuation}

Data valuation \citep[see][]{data_shapley} is the task of equitably quantifying the contribution of each data point to a model, via a data value $\phi_i(D, \mathcal{A}, V)$, where $D$ is the training dataset, $\mathcal{A}$ is a model and $V$ is a performance score, where $V(S)$ is the performance score of $\mathcal{A}$ when trained on $S \subseteq D$. 
\cite{Jiang23} propose a framework for benchmarking data valuation algorithms, concluding that no method uniformly outperforms others in every task. Therefore, in this work, we focus on a computationally efficient data valuation mechanism, which we propose to use for hardness characterization: \emph{KNN Shapley} \citep{knn_shapley}. 

KNN Shapleys correspond to an instance of the \emph{data Shapley} valuator \cite{data_shapley}. 
For the $i$-th training point in $D$, 
the data Shapley $
\phi_i$ takes the following form: 
\begin{equation}
    \phi_i = C \sum_{S \subseteq D\setminus\{i\}} {\binom{n-1}{|S|}}^{-1} \left[V(S \cup \{i\}) -V(S)\right],
\label{eq:datashap}
\end{equation}
where $C$ is an arbitrary constant. The valuator $\phi_i$ 
can be interpreted as weighted sum of all the possible ``marginal contributions'' of the datapoint. Exact evaluation of the data Shapleys is prohibitively expensive because of the need to retrain the model $2^{n-1}$ times (once per subset) for each observation in the training data. 
It is possible to approximate the data Shapleys by considering only a small number of subsets $S\subseteq D\setminus\{i\}$ in \eqref{eq:datashap}, obtained via Monte-Carlo sampling of random permutations. 
Unfortunately, even state-of-the-art approximations based on this idea, such as \textit{TMC-Shapley} \citep{data_shapley} or \textit{Beta-Shapley} \citep{beta_shapley} are computationally prohibitive for datasets with $n \gg 1000$, making them difficult to use in practice. 

An efficient method to calculate \emph{exact} data Shapleys for KNN classifiers in $\mathcal{O}\{n \log(n) \, n_{\mathrm{test}}\}$ complexity has been derived in \cite{knn_shapley}.
For training data $\{(\bm{x}_i, y_i)\}_{i=1}^n$ and test data $\{(\bm{x}_{\mathrm{test},i}, y_{\mathrm{test},i})\}_{i=1}^{n_{\mathrm{test}}}$, let 
$(\alpha_{j,1}, \ldots, \alpha_{j,n})$ be the indices of training data points, sorted in increasing order according to their Euclidean distance from the $j$-th test data point, for $j=1,\dots,n_{\mathrm{test}}$. The KNN Shapley for the $i$-th training data point is defined as 
$s_i=n_{\mathrm{test}}^{-1}\sum_{j=1}^{n_{\mathrm{test}}} s_{j,i}$, where $s_{j,i}$ is calculated recursively as follows:
\begin{align}
    & s_{j, \alpha_{j,n}} = \frac{1}{n}\mathds{1}_{y_{\mathrm{test},j}}\{y_{\alpha_{j,n}}\}, \label{knn_shaps} \\
    & s_{j, \alpha_{j,i}} = s_{j, \alpha_{j,i+1}} + \frac{\mathds{1}_{y_{\mathrm{test},j}}\{y_{\alpha_{j,i}}\} - \mathds{1}_{y_{\mathrm{test},j}}\{y_{\alpha_{j,i+1}}\}}{iK \cdot \min\{K, i\}^{-1}},
\end{align}
where $j=1,\dots,n_{\mathrm{test}}$, $i=1,\dots,n-1$, and $\mathds{1}_\cdot\{\cdot\}$ is the indicator function. 
\cite{scalability} found KNN Shapleys to be a valid alternative to data Shapleys in tasks such as data summarization or noisy labels detection, only when based on deep features extracted from image data by pre-trained architectures. A similar approach is unfortunately not possible on tabular data.
In this work we argue that KNN Shapleys 
can be used for hardness characterization on tabular data. Specifically, we argue that the \textit{lowest} KNN Shapleys identify the \textit{hardest} training points. 

\subsection{Synthetic data generation}\label{sec:sdg}
Synthetic data generators (SGDs) for tabular data have been actively researched in recent years \citep[see for example][]{Fonseca23}. 
One of the most popular SDGs used in practice is 
\textit{Synthetic Minority Over-sampling Technique} \citep[SMOTE;][]{smote}, 
which generates new data by random sampling along the segment connecting each data point to a randomly sampled neighbor. More recently, 
\textit{Conditional TGANs} (CTGAN), and \textit{Tabular VAE} (TVAE) were proposed in \cite{ctgan}. CTGAN consists in a \textit{Generative Adversarial Networks} \citep[GAN;][]{gan} with a conditional generator for tabular data which allows to generate data points conditional on specific values of the discrete features, for better handling imbalanced datasets, whereas TVAE is an adaptation for tabular data of a Variational Auto-Encoder \citep[VAE;][]{vae}. Diffusion and score-based methods have also been recently proposed in the literature \citep[see, for example][]{Kotelnikov23, Kim23, Zhang24}, as well as adversarial random forests \citep{Watson23}.

The objective of this work is \emph{not} to compare different SDGs, but rather to propose KNN Shapleys as a computationally efficient hardness characterizer, with the objective to extract ``hard'' data points to use in the training of \emph{any} synthetic data generator, making the procedure computationally more efficient without performance deterioration. Therefore, for the experiments in this work, we focus on two of the most popular and best performing SDGs in the literature (TVAE and CTGAN). However, we do emphasize that our procedure is applicable to \textit{any} SDG.  


%% file: sections/methods.tex
\section{KNN Shapleys as hardness characterizers for SDGs} \label{sec:methods}

In this work, we propose a simple pipeline: first, training data are ranked via a hardness characterizer. In particular, we propose to utilize a computationally efficient data valuator as hardness characterizer: KNN Shapleys, which can be calculated efficiently via \eqref{knn_shaps}. Second, we train a synthetic data generator using \emph{only the hardest points}, which provides a significant computational speedup over training with the entire training set. 
Our proposed pipeline is the following:
\begin{enumerate}
\item \textbf{KNN Shapley calculation} -- For a model $\mathcal A$ fitted on training data $D=\{(\bm{x}_i, y_i)\}_{i=1}^n$, calculate the KNN Shapleys $s_i,\ i=1,\dots,n$, based on a 
test set $\{(\bm{x}_{\mathrm{test},i}, y_{\mathrm{test},i})\}_{i=1}^{n_{\mathrm{test}}}$, using the recursion in \eqref{knn_shaps}.
\item \textbf{Ranking by hardness} -- Sort the KNN Shapley values in \textit{increasing order}, with points with lower values representing \emph{harder} examples. 
\item \textbf{Targeted augmentation} -- Given a synthetic data generator $\mathcal G$, perform data augmentation only on the $\tau\cdot100\%$ hardest points, based on the ranking of the KNN Shapleys, for a threshold $\tau\in(0,1]$. 
\end{enumerate}

\begin{figure}[t]
\centering
\includegraphics[width=\linewidth]{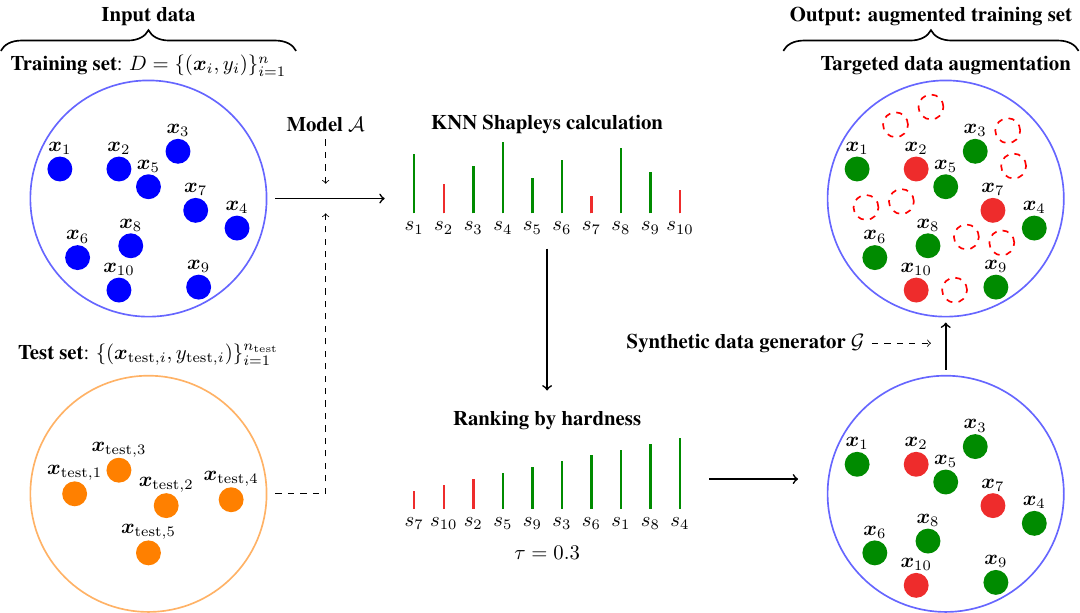}
\caption{Visual representation of the proposed targeted synthetic data generation pipeline.}
\label{fig:visual}
\end{figure}

By generating synthetic data for difficult examples only, we aim to improve the performance of the model $\mathcal A$ specifically on the most challenging parts of the data distribution. The proposed procedure is summarized visually in Figure~\ref{fig:visual}. 
This data processing approach could be related to the ideas of \emph{dataset distillation} \citep{Wang18} and \emph{dataset condensation} \citep{Zhao21}, commonly used in computer vision, aimed at finding compact representations of a large dataset via a small set of informative samples, for the purposes of data-efficient learning. 
The proposed approach could also be related to the \emph{data profiling} step in \cite{Hansen23}, which is used to divide training data points in different subsets (\textit{easy}, \textit{ambiguous} and \textit{hard}) via data valuation methods such as CleanLab \citep{Northcutt21} or Data-IQ \citep{dataiq}. Separate SDGs are fitted on each segment and the resulting synthetic data are combined, with the purpose of obtaining the best fully synthetic training dataset in terms of statistical fidelity. Therefore, the objective of \cite{Hansen23} is not data augmentation, but rather to generate a fully synthetic dataset which performs similarly to the original training data, to overcome data privacy issues. 
On the other hand, we \textit{exclusively focus} on data augmentation, with the objective of achieving the largest possible performance boost at a small computational cost: we propose to augment the hardest data points provided by the ranking of KNN Shapleys, \emph{combining} them with the original training dataset. 

%% file: sections/simulations.tex
\subsection{KNN Shapleys for hardness characterization: an example} \label{sec:toy_example}

In this section, we provide a mathematical intuition on the interpretation of KNN Shapleys as hardness charaterizers. 
 Consider a mixture of two univariate normals with unit variance, centred at $-1$ and $1$, such that $ p(x\mid y=0) = \mathcal N(x; -1,1)$ and $p(x\mid y=1) = \mathcal N(x; 1,1)$. Datapoints are drawn from each distribution with probability $1/2$.
Additionally, consider two points $(x_1,y_1)=(-1, 0)$ and $(x_2,y_2)=(1, 1)$, representative of each component of the mixture distribution. Consider a third point $(x_{\mathrm{train}}, y_{\mathrm{train}})$, with $y_{\mathrm{train}} = 0$ without loss of generality. Hence, the training set takes the form $D=\{(-1,0), (1,1), (x_{\mathrm{train}},0)\}$. As $x_{\mathrm{train}}$ increases, it is \textit{harder} to classify correctly under the true data distribution, since the ratio $p(x_{\mathrm{train}} \mid y=0)\ /\ p(x_{\mathrm{train}} \mid y=1)$ decreases monotonically (converging to zero in the limit), 
consequently increasing the probability of misclassification. 

\begin{figure}[t]
    \centering
    \begin{subfigure}{\linewidth}
    \centering
    \scalebox{0.85}{
    \begin{tabular}{cc|ccc}
        \toprule
        $x_{\mathrm{test}}$ & $y_{\mathrm{test}}$ & $s_{-1}$ & $s_{\mathrm{train}}$ & $s_{1}$ \\ 
        \midrule
        $(-\infty, -1.2)$ & $0$ & $1/2$ & $1/2$ & $0$ \\ 
        $(-\infty, -1/2)$ & $1$ & $-1/6$ & $-1/6$ & $1/3$ \\ 
        $(-1/2, 0)$ & $0$ & $1/2$ & $1/2$ & $0$ \\ 
        $(-1/2, 0)$ & $1$ & $-1/6$ & $-1/6$ & $1/3$ \\  
        $(0, 1/2)$ & $0$ & $1/3$ & $5/6$ & $-1/6$ \\ 
        $(0, 1/2)$ & $1$ & $0$ & $-1/2$ & $1/2$ \\ 
        $(1/2, +\infty)$ & $0$ & $1/3$ & $1/3$ & $-2/3$ \\ 
        $(1/2, +\infty)$ & $1$ & $0$ & $0$ & $1$ \\ 
        \bottomrule
        \end{tabular}
    }
    \caption{$1$NN Shapleys on the training data for $(x_{\mathrm{test}}, y_{\mathrm{test}})\in\mathbb R\times\{0,1\},\ x_{\mathrm{train}}=0$.}
    \label{tab:theor}
    \end{subfigure}
    
    \begin{subfigure}{\linewidth}
    \centering
    \includegraphics[width=\textwidth]{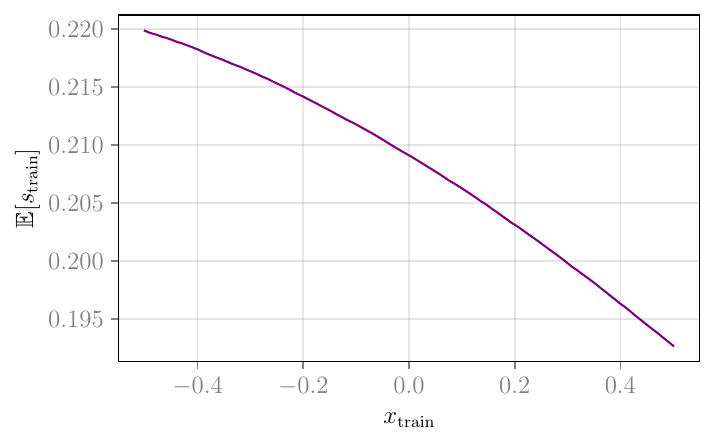}
    \caption{Expected value $\mathbb E(s_{\mathrm{train}})$ of the $1$NN Shapley for the data point $(x_{\mathrm{train}},0)$, for different values of $x_{\mathrm{train}}$.}
    \label{fig:theor}
    \end{subfigure}
    \centering
    \caption{Results for the experiment in Section~\ref{sec:toy_example}, based on a mixture of two normals with unit variance and means $-1$ and $1$, and training data $\{(-1,0),(1,1),(x_{\mathrm{train}},0)\}$.}
    \label{fig:gaussians}
\end{figure}

Any test data point $(x_{\mathrm{test}}, y_{\mathrm{test}})\in\mathbb R\times\{0,1\}$ induces a ranking of the training data based on the distances $\vert x_{\mathrm{test}}-x \vert,\ x\in\{x_1,x_2,x_{\mathrm{train}}\}$.
Assuming $x_{\mathrm{train}} = 0$, the $1$NN Shapley 
value $s_{\mathrm{train}}$ for the point $(x_{\mathrm{train}}, y_{\mathrm{train}})=(0,0)$ can be calculated explicitly for all the possible values of $(x_{\mathrm{test}}, y_{\mathrm{test}})$. The results are reported in Table~\ref{tab:theor}.
Since the data distribution is known, the expected value $\mathbb{E}(s_{\mathrm{train}})$ of the $1$NN Shapley for $(x_{\mathrm{train}}, y_{\mathrm{train}})=(0,0)$ can also be explicitly calculated: 
\begin{multline}
    \mathbb{E}(s_{\mathrm{train}}) = \int_{-\infty}^{+\infty} s_{\mathrm{train}}\, p(x_{\mathrm{test}})\ \mathrm{d}x_{\mathrm{test}} 
    \\ = \frac{1}{2}\sum_{y\in\{0,1\}} \int_{-\infty}^{+\infty} s_{\mathrm{train}}\, p(x_{\mathrm{test}}\,|\,y)\ \mathrm{d}x_{\mathrm{test}}
    \approx 0.209.
    \label{eq:integral}
\end{multline}
The value of $s_{\mathrm{train}}$ depends on $(x_{\mathrm{test}}, y_{\mathrm{test}})$, as detailed in Table~\ref{tab:theor} and \eqref{eq:datashap}.
The same procedure as \eqref{eq:integral} can be repeated for any $x_{\mathrm{train}}\in\mathbb R$, with results displayed in Figure \ref{fig:theor}. As expected, the $1$NN Shapley decreases in the direction of increasing hardness, suggesting an association between low KNN Shapleys and hard regions of feature space.

%% file: sections/results.tex
\section{Results}\label{sec:results}
In this section, the performance of KNN Shapleys as hardness characterizer is first compared to the most common methods in the literature via  benchmarks on tabular data. Second, we demonstrate the pipeline proposed in Section~\ref{sec:methods} on a simulated dataset. Finally, KNN Shapleys are calculated on a number of real-world tabular datasets to identify the hardest points, which are later used to train and compare synthetic data generators.

\subsection{Benchmark study}\label{sec:benchmark}

\begin{table*}[ht]
\centering
\scalebox{0.75}{
\begin{tabular}{l|cccc|cccc|cccc|cccc}
\toprule
 & \multicolumn{16}{c}{\textbf{Dataset}} \\
\cline{2-17}\\[-1em]
& \multicolumn{4}{c|}{\texttt{diabetes}} & \multicolumn{4}{c|}{\texttt{cover}} & \multicolumn{4}{c|}{\texttt{eye}} & \multicolumn{4}{c}{\texttt{jannis}} \\
\cline{2-17}\\[-1em]
\textbf{Hardness} & \multicolumn{4}{c|}{Proportion of perturbed points} & \multicolumn{4}{c|}{Proportion of perturbed points} & \multicolumn{4}{c|}{Proportion of perturbed points} & \multicolumn{4}{c}{Proportion of perturbed points} \\
\textbf{characterizer} & 0.05 & 0.1 & 0.15 & 0.2 & 0.05 & 0.1 & 0.15 & 0.2 & 0.05 & 0.1 & 0.15 & 0.2 & 0.05 & 0.1 & 0.15 & 0.2 \\
\midrule
Agreement & 0.951 & 0.927 & 0.902 & 0.877 & 0.954 & 0.932 & 0.908 & 0.885 & 0.951 & 0.927 & 0.903 & 0.878 & 0.951 & 0.927 & 0.903 & 0.879 \\
AUM & 0.956 & 0.934 & 0.912 & 0.890 & 0.965 & 0.947 & 0.928 & 0.908 & 0.953 & 0.932 & 0.909 & 0.885 & 0.961 & 0.942 & 0.922 & 0.902 \\
CleanLab & 0.956 & 0.933 & 0.911 & 0.888 & 0.967 & 0.950 & 0.930 & 0.911 & 0.954 & 0.933 & 0.911 & 0.887 & 0.961 & 0.942 & 0.922 & 0.902 \\
Data-IQ & 0.955 & 0.934 & 0.912 & 0.889 & 0.965 & 0.946 & 0.926 & 0.905 & 0.954 & 0.933 & 0.911 & 0.887 & 0.961 & 0.942 & 0.922 & 0.902 \\
DataMaps & 0.955 & 0.934 & 0.912 & 0.889 & 0.965 & 0.946 & 0.926 & 0.905 & 0.954 & 0.933 & 0.911 & 0.887 & 0.961 & 0.942 & 0.922 & 0.902 \\
Detector & 0.950 & 0.925 & 0.900 & 0.875 & 0.945 & 0.920 & 0.894 & 0.869 & 0.950 & 0.925 & 0.900 & 0.875 & 0.946 & 0.920 & 0.895 & 0.870 \\
EL2N & 0.942 & 0.914 & 0.886 & 0.859 & 0.922 & 0.890 & 0.861 & 0.832 & 0.945 & 0.915 & 0.886 & 0.860 & 0.926 & 0.894 & 0.863 & 0.833 \\
Forgetting scores & 0.951 & 0.925 & 0.900 & 0.874 & 0.951 & 0.927 & 0.902 & 0.877 & 0.951 & 0.926 & 0.901 & 0.875 & 0.951 & 0.927 & 0.902 & 0.877 \\
GraNd & 0.932 & 0.900 & 0.869 & 0.839 & 0.942 & 0.916 & 0.891 & 0.867 & 0.946 & 0.920 & 0.893 & 0.867 & 0.935 & 0.907 & 0.879 & 0.852 \\
KNN-Shapleys & 0.957 & 0.924 & 0.902 & 0.877 & 0.961 & 0.940 & 0.919 & 0.898 & 0.950 & 0.924 & 0.900 & 0.875 & 0.951 & 0.927 & 0.902 & 0.875 \\
Prototypicality & 0.953 & 0.927 & 0.904 & 0.880 & 0.968 & 0.950 & 0.929 & 0.909 & 0.953 & 0.932 & 0.909 & 0.885 & 0.959 & 0.939 & 0.918 & 0.897 \\
Sample-loss & 0.941 & 0.913 & 0.885 & 0.857 & 0.922 & 0.891 & 0.863 & 0.836 & 0.944 & 0.915 & 0.885 & 0.858 & 0.925 & 0.893 & 0.862 & 0.833 \\
VoG & 0.963 & 0.942 & 0.922 & 0.900 & 0.964 & 0.944 & 0.923 & 0.903 & 0.950 & 0.925 & 0.899 & 0.874 & 0.957 & 0.934 & 0.912 & 0.889 \\
\bottomrule
\end{tabular}
}
\caption{AUPRC scores for 13 different hardness characterizers (including KNN Shapleys) on the \texttt{diabetes}, \texttt{cover}, \texttt{eyes} and \texttt{jannis} OpenML datasets, for different proportions of perturbed points.}
\label{tab:benchmarks}
\end{table*}

We demonstrate that KNN Shapleys have comparable performance to alternative hardness characterizers on tabular data. 
We provide a comparison with the results in the benchmarking study from \cite{hard_cat}, where the problem of hardness characterization is approached quantitatively by comparing existing methods on how confidently they can identify different kinds of hard datapoints on a variety of OpenML datasets \citep{OpenML}. Their toolkit focuses on benchmarking four OpenML tabular datasets: \texttt{diabetes}, \texttt{cover}, \texttt{eye} and \texttt{jannis}. 
We compare KNN Shapleys to the following hardness characterization methods identified in \cite{hard_cat}: \textit{Agreement} \citep{Carlini19}, \textit{AUM} \citep{Pleiss20}, \textit{CleanLab} \citep{Northcutt21}, \textit{Data-IQ} \citep{dataiq}, \textit{DataMaps} \citep{Sway20}, \textit{Detector} \citep{Jia23}, \textit{EL2N} \citep{Paul21}, \textit{Forgetting scores} \citep{Toneva19}, \textit{GraNd} \citep{Paul21}, \textit{Prototipicality} \citep{Sorsh22}, \textit{Sample-loss} \citep{Xia22}, and \textit{VoG} \citep{Agarwal22}. In particular, for each dataset we perturb a chosen proportion $p$ of datapoints according to one of three hardness types (\textit{mislabeling}, \textit{out-of-distribution}, or \textit{atypical}; \textit{cf.}~Section~\ref{sec:hard_cat}), with $p \in \{0.05, 0.1, 0.15, 0.2\}$. We use the \textit{Area Under the Precision Recall Curve} (AUPRC) of an MLP as the performance metric in order to facilitate direct comparisons, since most hardness characterizers analyzed in \cite{hard_cat} are devised for neural networks and do not support XGBoost. 
  The results are plotted in Table~\ref{tab:benchmarks}. 
  Scores are averaged across three independent runs per hardness type and across hardness types. 


Beyond the justifications described in Section~\ref{sec:methods}, the results of the benchmarking study suggest that low KNN Shapleys are solid identifiers of hardness, with performances comparable with the most common techniques in literature. It must be emphasized that we do not claim that KNN Shapleys outperform other hardness characterizers, but we point out that the proposed scores achieve similar AUPRC to state-of-the-art methods, \emph{at a fraction of their computational cost}. In particular, the main advantage lies in the fact that KNN Shapleys do not require training of \textit{any} model, and they only require a calculation of a distance matrix. On the other hand, all the other methods in Table~\ref{tab:benchmarks} require training of a model for calculation of the hardness characterization scores. This makes KNN Shapleys computationally efficient and model-agnostic, whereas most other methods require multiple model refitting procedures, rely on the training dynamics of neural networks, and it is not straighforward to extend them to gradient boosting for tabular data. 

\subsection{Simulation study} \label{sec:sim_ex}

\begin{figure}[t]
    \centering
        \begin{subfigure}{0.475\textwidth}
        \centering
            \caption{Scatter of training data (left) and hardest $5\%$ (right).}
            \label{fig:scatter}
            \includegraphics[width=.475\textwidth]{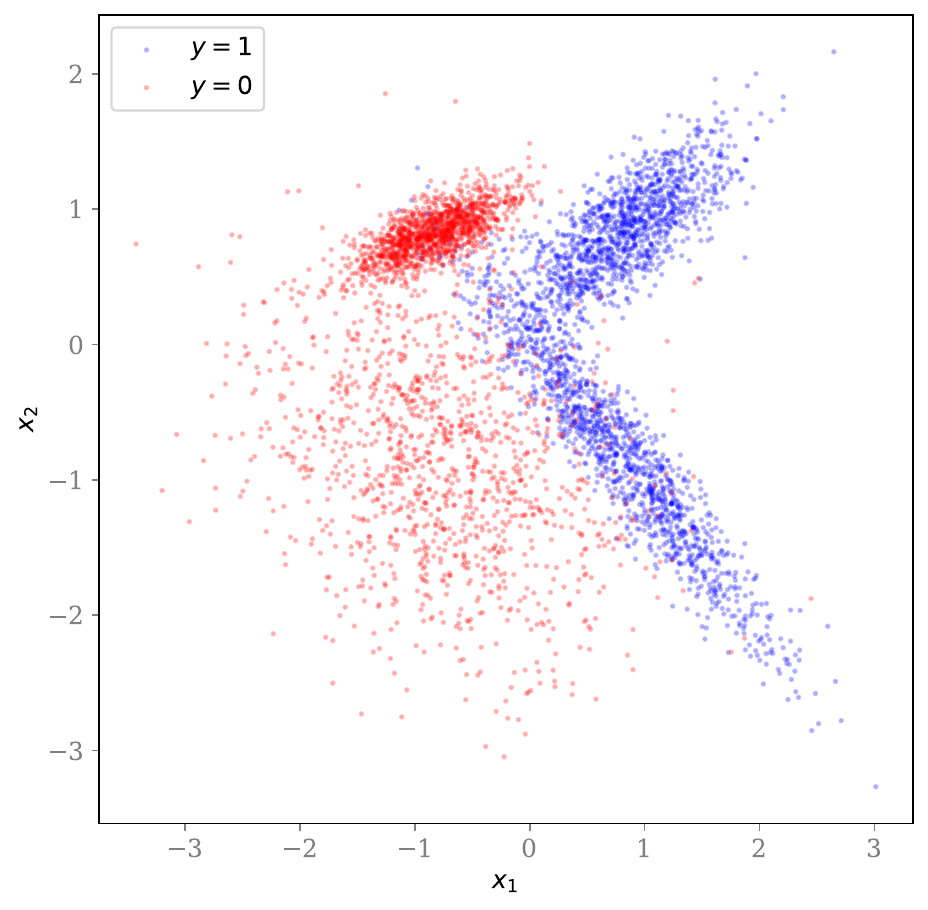}
            \includegraphics[width=.475\textwidth]{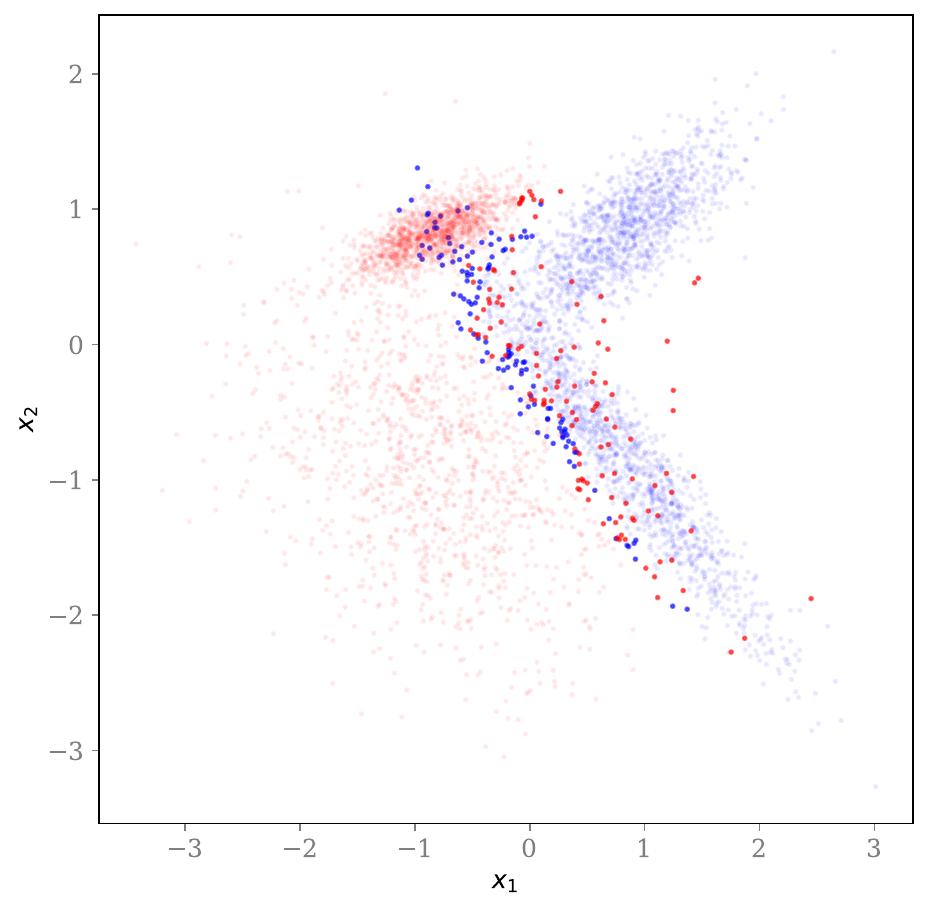}
        \end{subfigure}
        \begin{subfigure}{0.475\textwidth}
        \centering
        \caption{Scores for TVAE after augmentation with $95\%$ CIs estimated by generating synthetic data 30 times.}
        \label{fig:comp}
        \includegraphics[width=0.95\textwidth]{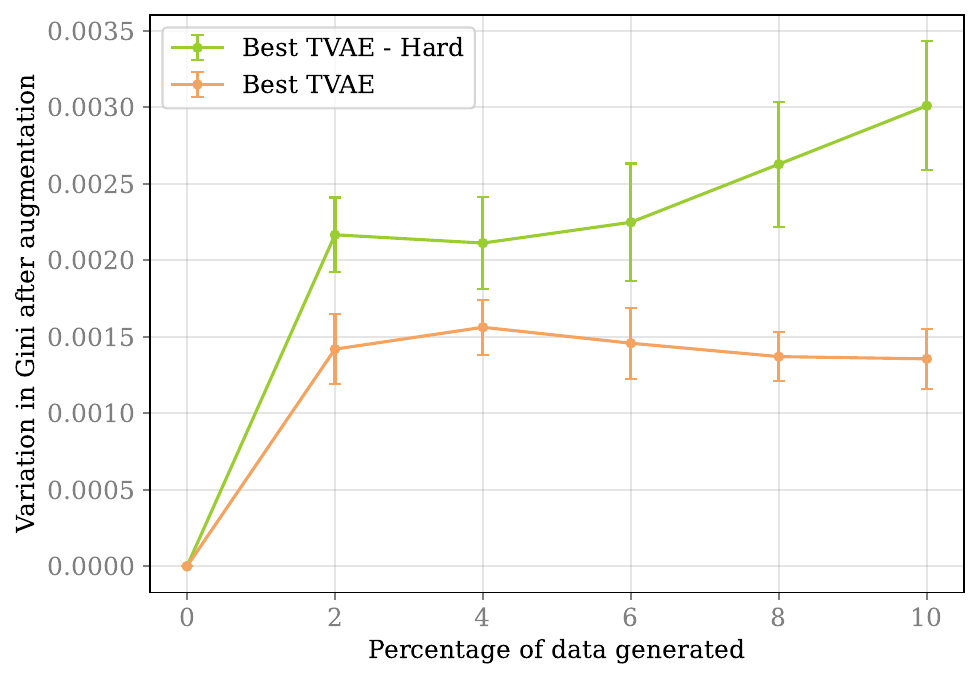}
        \end{subfigure}
    \caption{Gini scores under different synthetic data augmentation regimes on the bivariate Gaussian simulated data, with scatterplots of observations and 5\% hardest data points.}
    \label{fig:tuning_sim}
\end{figure}

To demonstrate the performance of the procedure proposed in Section~\ref{sec:methods}, we run the hardness characterization and data augmentation pipeline on simulated data. Specifically, we consider four bivariate normal distributions, assigning to two of them label $y = 0$, and $y = 1$ to the remaining ones. 
We draw $n_{\mathrm{train}} = 5\,000$ training points, $n_{\mathrm{valid}} = 2\,500$ observations for model validation and lastly $n_{\mathrm{test}} = 2\,500$ points to calculate $5$NN Shapleys. The training dataset can be visualized in Figure \ref{fig:scatter} on the left, while the $5\%$ hardest points according to $5$NN Shapleys are shown on the right: notice that they are concentrated around the decision boundary, with most of them falling on the ``wrong'' side. This is a common issue with tabular data, referred to as \textit{outcome heterogeneity} \citep{dataiq}.

We proceed by tuning TVAE using the GPEI algorithm (described in Appendix~\ref{sec:ap_training}) both when augmenting the hardest $\tau\cdot100\%$ by $100\%$ and when augmenting the entire dataset by $\tau\cdot100\%$, for $\tau\in\mathbb R_+$. 
Information on the tuning process can be found in Appendix~\ref{sec:a2}.
Performance is measured via the normalized Gini coefficient $2 \cdot \mathrm{AUCROC} - 1$, with $\mathrm{AUCROC}$ denoting the \textit{Area Under the Receiver Operating Characteristic} curve. 
Figure~\ref{fig:comp} displays the results for different values of $\tau$. 
The targeted approach results in a larger performance improvement, both for the best attempt and on average across different sets of hyperparameters.  
Hard points augmentation consistently outperforms the non-targeted approach and improves as more data is added.

\subsection{Case study: American Express data}\label{sec:amex_res}

The main real-world benchmark for demonstrating the methods discussed in this work is the American Express credit default prediction dataset \citep{amex}, referred to as the \textit{Amex dataset}. The dataset consists of observations for $n = 458\,913$ customers. For each of them, we consider 188 features available at the latest credit card statement date, in addition to a binary label $y_i \in \{0, 1\}$ which specifies whether or not the customer was able to repay their credit card debt within 120 days of the statement.
The $n = 458\,913$ available observations are split into $n_{\mathrm{valid}} = 50\,000$ datapoints used to validate model performance on unseen data, $n_{\mathrm{test}} = 50\,000$ observations used to calculate KNN Shapleys, and the remaining datapoints are used for training. The splits are performed randomly stratifying on the target, so that the same ratio of defaulters and non-defaulters is maintained across datasets. Performance is measured via the normalized Gini coefficient. 
XGBoost was found to have the best performance \textit{on its own} 
and is thus chosen as baseline for this study. 
Through experimenting it was found that setting the maximum tree depth to 4, the learning rate to 0.05, the subsampling ratio of datapoints to 0.8 and of features to 0.6, gives a validation Gini of $0.91986$ in under 2 minutes of training time.

\paragraph{Hardness characterization}

\begin{figure}[t]
    \centering
    \includegraphics[width=0.925\linewidth]{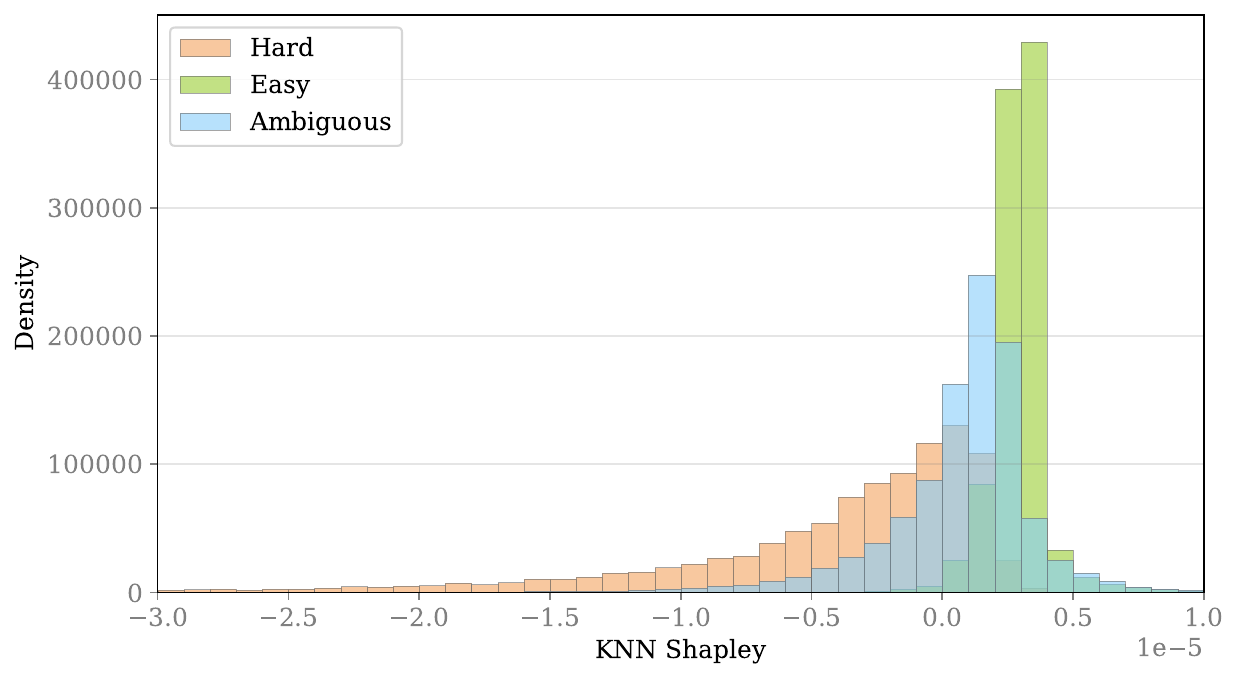}
    \caption{Distribution of the 100NN data Shapley scores for different Data-IQ tags.}
    \label{fig:knn_by_dataiq}
\end{figure}

\begin{figure}[t]
    \centering
    \includegraphics[width=0.925\linewidth]{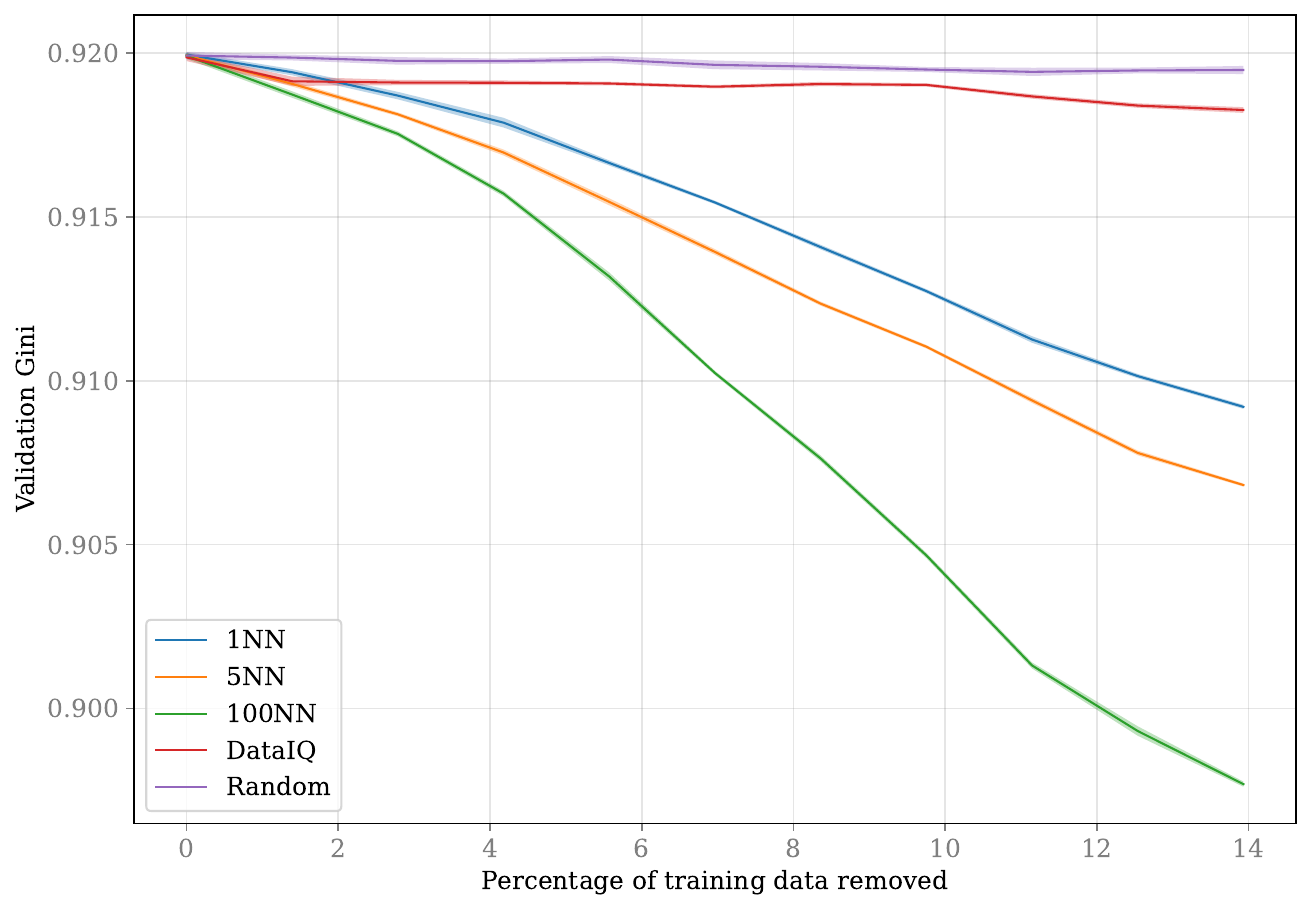}
    \caption{Validation Gini after removing the hardest points with KNN Shapleys, Data-IQ, and randomly, 
    with $95\%$ CIs.}
    \label{fig:comp_hard}
\end{figure}

KNN Shapleys are computed for $K \in \{1, 5, 100\}$ under the assumption that the $n_{\mathrm{test}}$ available out-of-sample data points are representative of the entire data distribution.
Figure \ref{fig:knn_by_dataiq} displays the distribution of 100NN Shapleys separated by Data-IQ's tag: hard points as per Data-IQ exhibit low 100NN Shapleys, with very good adherence for the hardest points. In addition, the left tail of the ambiguous points lies in the region of low 100NN Shapleys and would thus be recognized as hard by our novel characterizer.
To choose the best $K$, validation Gini is monitored as we re-train XGBoost after gradually removing the hardest datapoints. These points are expected to be the most valuable for XGBoost and thus the best hardness characterizer should be the one displaying the fastest drop in validation performance. 100NN Shapleys outperform both other choices of $K$ and Data-IQ (\textit{cf.} Figure \ref{fig:comp_hard}). This comparison highlights an issue with Data-IQ: after a sharp drop when the hard points are removed, performance decreases at a rate comparable to random as the most ambiguous points are removed. \cite{dataiq} argue that the removal of ambiguous points should actually make the model more robust, as aleatoric uncertainty is data-dependent and cannot be reduced unless additional features are collected. This is however not the case for the Amex dataset, and beyond the small amount of hard points identified, Data-IQ does not provide guidance on the next most valuable points. 

\paragraph{Synthetic data generation}
The $10\%$ hardest datapoints are chosen as training set for synthetic data generators, with the purpose of establishing whether augmenting only the hardest datapoints makes XGBoost more robust than non-targeted data augmentation. The choice of a hard threshold is imposed by the expensive computational cost of fine-tuning neural networks, and $10\%$ is justified as it is just before the elbow region where the decrease in validation performance when removing hard data slows down (\textit{cf.} Figure \ref{fig:comp_hard}). 
Details about the training of the SDGs are given in Appendix~\ref{sec:ap_training}.

For models trained on the $10\%$ hardest points, we augment by $100\%$, while for models trained using the entire dataset we augment by $10\%$, thus guaranteeing the same amount of synthetic samples across all experiments. Notice that for each attempt synthetic data has to be generated multiple times with different seeds to mitigate randomness and quantify uncertainty on the performance metric. 

\begin{figure}[t]
    \centering
        \begin{subfigure}{0.475\textwidth}
        \centering
            \caption{Variation in Gini after augmentation with $95\%$ CIs.}
            \label{tab:res}
            \scalebox{0.75}{
                \begin{tabular}{ccc}
                    \toprule
                    & hardest $10\%$ by $100\%$ & full data by $10\%$ \\
                    \midrule
                    SMOTE & $[-0.00122, -0.00109]$ & $[0.000116, 0.000192]$ \\
                    Best TVAE & $[0.000366, 0.000431]$ & $[0.000232, 0.000278]$ \\
                    Best CTGAN & $[0.000193, 0.000312]$ & $[0.000135, 0.000237]$ \\
                    \bottomrule
                \end{tabular}
            }
        \end{subfigure}
        \begin{subfigure}{0.475\textwidth}
        \centering
        \caption{Scores for TVAE and TVAE on hard data points after $5\%, 10\%, 15\%, 20\%$ augmentation, with $95\%$ confidence intervals estimated by generating synthetic data 10 times.}
        \label{fig:tvae_repeat}
        \includegraphics[width=0.95\textwidth]{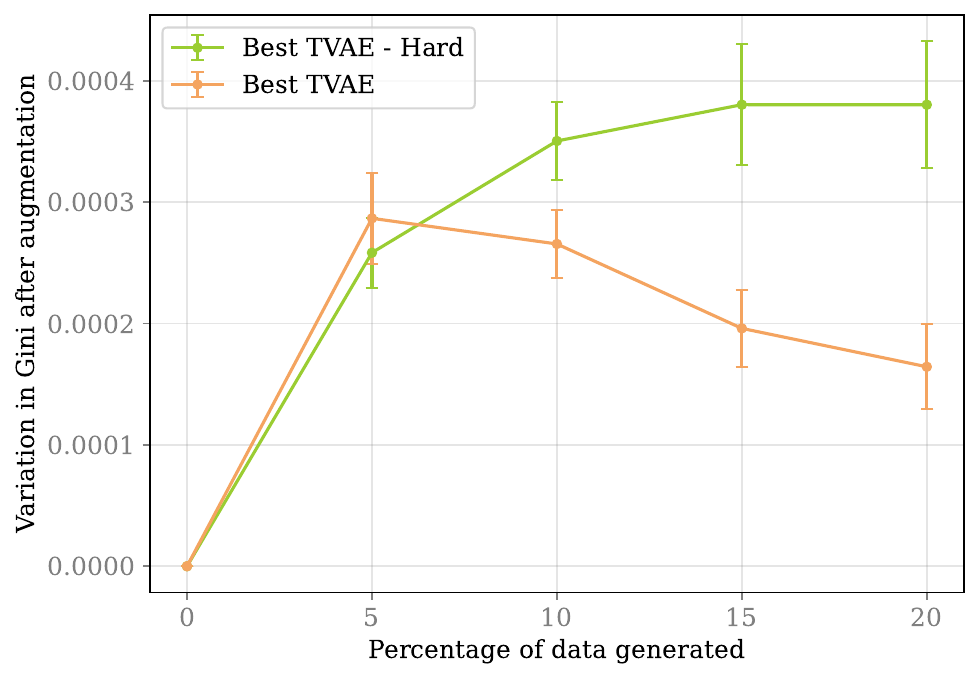}
        \end{subfigure}
    \caption{Gini scores under different synthetic data augmentation regimes on the Amex dataset.}
    \label{fig:tuning}
\end{figure}

The best scores are in both cases achieved by TVAE, and augmenting only the hardest points generally leads to a more significant improvement in validation performance, both for the best attempts and on average across different sets of hyperparameters. Information on the tuning grid and best hyperparameters are detailed in Appendix~\ref{sec:a1}.


Table \ref{tab:res} reports the performance variations for the best TVAE, CTGAN and SMOTE: the latter does not work for the hardest points and is outperformed by TVAE for general augmentation. The robustness of these results for the best attempts can be verified by augmenting by different amounts ($5\%, 10\%, 15\%, 20\%$), with results displayed in Figure~\ref{fig:tvae_repeat}: generating more hard points further increases performance with diminishing returns after $15\%$, while for non-targeted augmentation, the performance actually worsens as more data is added due to the extra noise. Finally, to quantify the magnitude of this improvement, adding back the $n_{\mathrm{test}} = 50\,000$ observations used to compute KNN Shapleys into training data improves validation performance by $\approx 0.000326$, which is less than the improvement obtained from augmentation of the hard points via TVAE.  

It could be argued that the improvement in validation performance achieved via data augmentation, although larger for models trained on only the hardest datapoints, is very small in absolute value. 
It must be remarked that similar improvements have only been achieved in the original Kaggle competition by generating thousands of features and building large ensemble models. Additionally, XGBoost already classifies over $90\%$ of validation data correctly, and any further improvement on this already highly performing model may have business value in practice.

\subsection{Additional results} \label{sec:add_res}

\begin{table*}[!t]
    \centering
    \begin{tabular}{rllll}
        \toprule 
        \bfseries Dataset & \multicolumn{1}{
        p{3cm}}{ \bfseries Augmenting hardest 5\% by 200\%} & \multicolumn{1}{p{3cm}}{\bfseries Augmenting hardest 10\% by 100\%} & \multicolumn{1}{p{3cm}}
        {\bfseries Augmenting hardest 20\% by 50\%} & \multicolumn{1}{p{3cm}}
        {\bfseries Augmenting by 10\% (non targeted)} \\
        \midrule 
        \texttt{diabetes} & 0.00084 & 0.00050 & \bfseries 0.00193 & 0.00159 \\
        \texttt{credit\_card} & 0.00429 & \bfseries 0.00528 & 0.00173 & 0.00303 \\
        \texttt{blobs} & \bfseries 0.00049 & -0.00130 & -0.00127 & -0.00255 \\
        \texttt{support} & 0.00229 & 0.00379 & \bfseries 0.00716 & 0.00592 \\
        \bottomrule 
    \end{tabular}
        \caption{Variation in Gini after 10\% data augmentation for targeted and non-targeted approaches on four different datasets.} \label{tab:supp-data}
\end{table*}

To extend the results beyond the case study on the Amex dataset, and further verify the robustness of the proposed pipeline, the same hardness characterization and data augmentation pipeline are applied to other common binary classification datasets from the UC Irvine (UCI) machine learning repository. Similarly to the Amex dataset, hyperparameter tuning of CTGAN and TVAE is performed at the synthetic data generation stage, and the randomness of both XGBoost training and data augmentation is mitigated by taking the mean over multiple runs. In addition, the smaller size of these datasets allows to experiment with different thresholds for hardness characterization, specifically 5\%, 10\% and 20\%. The results are reported in Table~\ref{tab:supp-data}. The results demonstrated that targeted augmentation outperforms the non-targeted approach for all datasets, with the choice of the threshold $\tau$ proving critical.

%% file: sections/discussion.tex
\section{Conclusion and discussion} \label{sec:discussion}

In this work, we verified empirically 
that, when it comes to data augmentation, focusing only on the most challenging points for the model can be beneficial, both in terms of performance improvement and computational efficiency. To achieve this result, we first devised a novel hardness characterizer based on KNN Shapleys, capable of achieving performances comparable to state-of-the-art methods in hardness detection benchmarks on tabular data. Then, we performed a complex fine-tuning routine on synthesizers trained either on the entire dataset or on the hardest points only, allowing to rigorously establish which approach leads to the largest performance improvement.

Methodologically, this work aims at bridging the gap between learning-based hardness characterization and game theoretic data valuation. The benchmarks reported in Section \ref{sec:benchmark} represent the first quantitative comparison between a Shapley-based evaluator and existing hardness characterizers. While research on the two topics has often dealt with similar problems, such as mislabeling or data summarization \citep{grand, scalability}, the two paths have never crossed.

Practically, this work attempts to change the perspective on data summarization: while existing literature has mostly focused on pruning the least valuable points from training data \citep{dataiq, dataoob}, here we propose augmenting the most valuable ones. We demonstrate that not all datapoints are created equal, and some play a more significant role in determining the predictive power of the final model: high-value data dictates which points we should either collect in larger amounts or, when this is not possible, synthetically generate.



%% file: sections/reproducibility.tex
\subsubsection*{Details on reproducibility}

The Amex dataset is publicly available on the public data repository \href{https://www.kaggle.com/competitions/amex-default-prediction/data}{Kaggle}. This work uses the denoised version by user \textit{raddar}, which is publicly available on Kaggle in \href{https://www.kaggle.com/datasets/raddar/amex-data-integer-dtypes-parquet-format?select=train.parquet}{\texttt{parquet} format}.
The code to reproduce the results in this work is available in the GitHub repository \if1\blind{\href{https://github.com/tommaso-ferracci/Targeted_Augmentation_Amex}{\texttt{tommaso-ferracci/Targeted\_Augmentation\_\allowbreak Amex}}}\fi\if0\blind{\texttt{anonymised\_link}}\fi. In particular, code to reproduce the hardness characterization and data augmentation pipeline on both the Amex dataset and the benchmark datasets is available in the \texttt{analyses/} folder. The \texttt{outputs/} folder contains, in addition to figures and results, the model weights for the best synthesizers. Every method from the \texttt{ctgan} and \texttt{sdv} packages uses forked versions (\if1\blind{\href{https://github.com/tommaso-ferracci/CTGAN}{\texttt{tommaso-ferracci/CTGAN}}}\fi\if0\blind{\texttt{anonymised\_link}}\fi\, and \if1\blind{\href{https://github.com/tommaso-ferracci/SDV}{\texttt{tommaso-ferracci/SDV}}}\fi\if0\blind{\texttt{anonymised\_link}}\fi) with custom early stopping. Detailed instructions on how to install these dependencies are available in the \texttt{Makefile}. Details around 
training are also reported in Appendices~\ref{sec:ap_training}, \ref{sec:a1} and~\ref{sec:a2}. 

%% file: sections/appendix.tex
\subsection{Training CTGAN and TVAE -- Amex dataset} \label{sec:ap_training}

\begin{figure}[t]
    \centering
    \begin{subfigure}{0.475\textwidth}
        \caption{CTGAN}
        \label{fig:ctgan_hard_loss}
        \includegraphics[width=\textwidth]{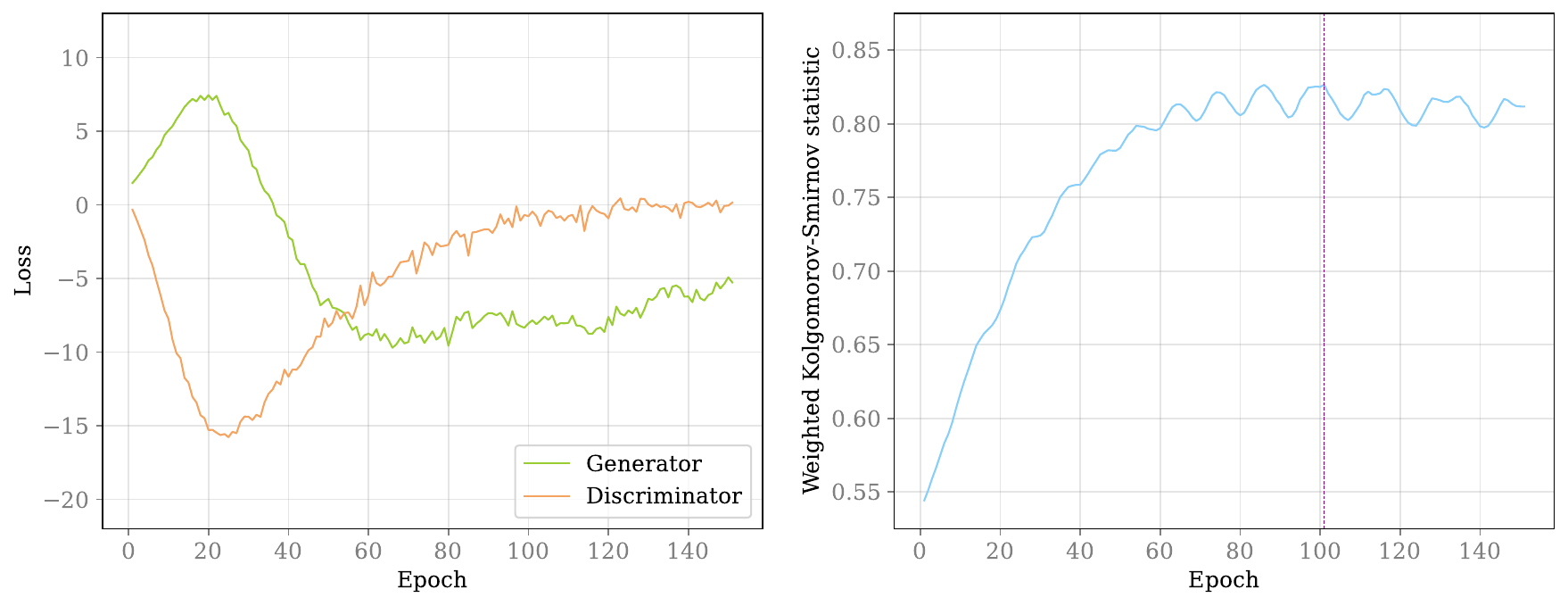}
    \end{subfigure}
    \hspace*{0.025\textwidth}
    \begin{subfigure}{0.475\textwidth}
        \caption{TVAE}
        \label{fig:tvae_hard_loss}
        \includegraphics[width=\textwidth]{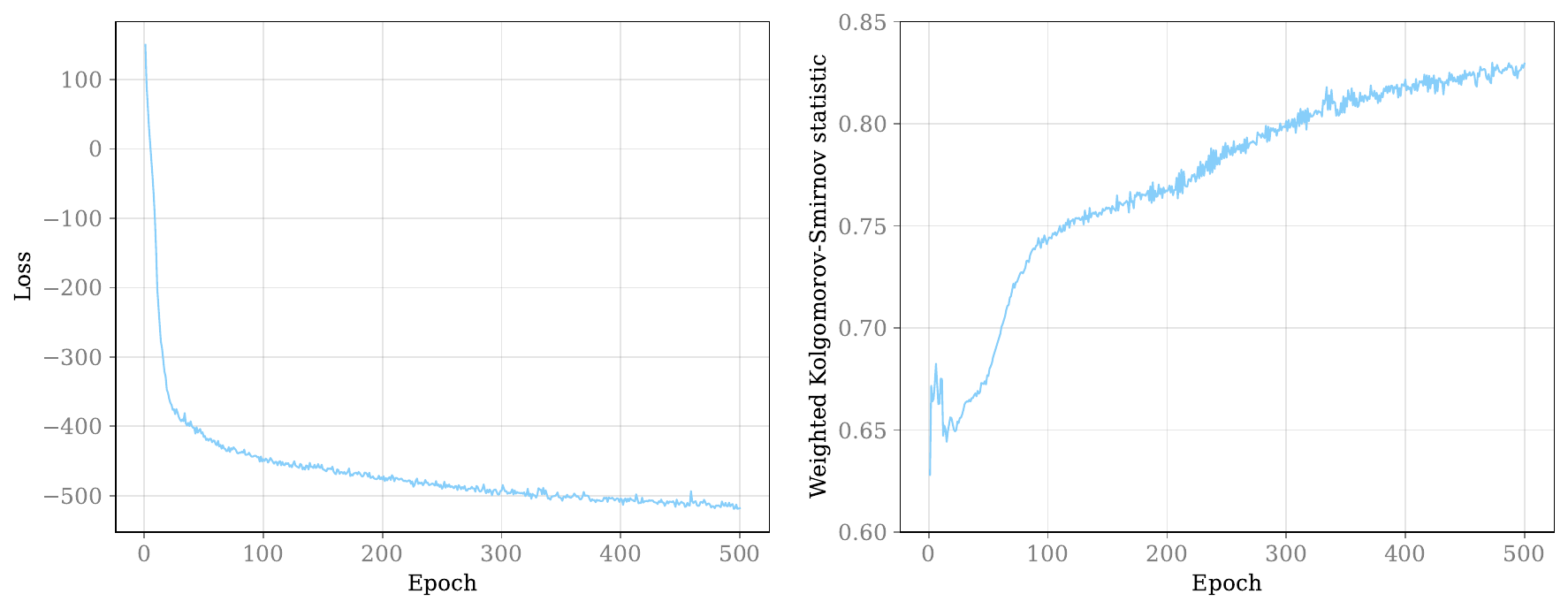}
    \end{subfigure}
    \caption{Loss and KS statistics for training CTGAN and TVAE.}
\end{figure}

In the following, we provide details about CTGAN and TVAE SDGs on the Amex dataset.
A common issue when training GANs is the instability of both generator and discriminator losses. More specifically, the two typically move in opposite ways with oscillatory patterns, making it difficult to decide when to stop training. For this reason, we introduce a novel early stopping condition which tracks epoch after epoch a weighted average of the \textit{Kolgomorov-Smirnov} (KS) statistic between real and synthetic data across all individual features. In particular, we choose to use as weights the feature importances, in order to focus more on the features relevant to XGBoost. The patience is set to $50$ epochs with a maximum number of epochs of $500$. If the early stopping condition is triggered, the model is reverted to the best epoch. A large batch size of $10\,000$ is chosen to limit the number of updates per epoch and guarantee a smoother training process. Both the generator and discriminator are trained using the Adam optimizer \citep{adam}, setting the learning rate to $2\cdot10^{-4}$, weight decay to $10^{-6}$, and the momentum parameters to $\beta_1 = 0.5$ and $\beta_2 = 0.9$.
Figure \ref{fig:ctgan_hard_loss} shows the losses and the statistic epoch by epoch: we can see that the losses move against each other and then more or less converge once the model cannot further improve. We notice oscillatory patterns in the tracked statistic, symptomatic of the training dynamics of the generator and discriminator pair, and the early stopping condition kicking in after around 100 epochs when the weighted KS statistic peaks at $0.83$. 

Training of the VAE relies on the Adam optimizer with a learning rate of $10^{-3}$, weight decay of $10^{-5}$, $\beta_1$ of $0.9$ and $\beta_2$ of $0.999$. 
To avoid overfitting of training data and the subsequent generation of exact replicas of training points, we stop training once the maximum number of epochs is reached. 
The following routine integrating both model selection and hyperparameter tuning is carried out separately for models trained on only the hardest points and models trained using the entire dataset:
\begin{itemize}
    \item \textbf{Model selection}: after training one instance of TVAE and CTGAN to gather initial data, for each candidate model $i$ the \textit{Upper Confidence Bound 1} (UCB1) score is calculated:
    $$\mathrm{UCB1}(i) = \bar{\psi}_i + \sqrt{\frac{2 \ln t}{n_i}},$$ where $\bar{\psi}_i$ is the average score of model $i$, $t$ is the total number of models trained, and $n_i$ is the number of times model $i$ is selected. We select the model with the highest UCB1 score, which results from a balance of exploration (low $n_i$) and exploitation (high $\bar{\psi}_i$). 
    \item \textbf{Hyperparameter tuning}: once model $i$ has been proposed, the set of hyperparameters to try is chosen via \textit{Gaussian Process Expected Improvement} (GPEI): a \textit{Gaussian Process} (GP) is fitted to $\{(\mathbf{h}_j, \psi_j)\}_{j=1}^{n_i}$, with $\mathbf{h}_j$ hyperparameters and $\psi_j$ corresponding score, and then the new set to try is chosen to maximize over $\mathbf{h}$ the \textit{Expected Improvement} $\mathrm{EI}(\mathbf{h}) = \mathbb{E}[\max(0, f(\mathbf{h}) - f(\mathbf{h}^*))]$, with $f$ mean function of the fitted GP and $\mathbf{h}^*$ the best set so far.
\end{itemize}

The score $\psi$ over which the optimization is carried out, is the variation in validation Gini after training XGBoost on the augmented dataset. 

\subsection{Hyperparameters - Amex dataset}\label{sec:a1}

\begin{figure}[!ht]
\begin{minipage}[b]{0.45\textwidth}
\centering
\scalebox{0.85}{
\begin{tabular}{lcccc}
\toprule
Hyperparameter & LB & UB & Default & Best \\ \midrule
Embedding dim. & 32 & 512 & 64 & \textbf{64} \\ 
Encoder dim. 1 & 32 & 512 & 128 & \textbf{128} \\ 
Encoder dim. 2 & 32 & 512 & 128 & \textbf{128} \\ 
Decoder dim. 1 & 32 & 512 & 128 & \textbf{128} \\ 
Decoder dim. 2 & 32 & 512 & 128 & \textbf{128} \\ \bottomrule
\end{tabular}
}
\captionof{table}{TVAE, hard 10\%: tuning setup.}
\label{tab:tvae_hard_amex}
\end{minipage}
\begin{minipage}[b]{0.55\textwidth}
\centering
\includegraphics[width=.9\textwidth]{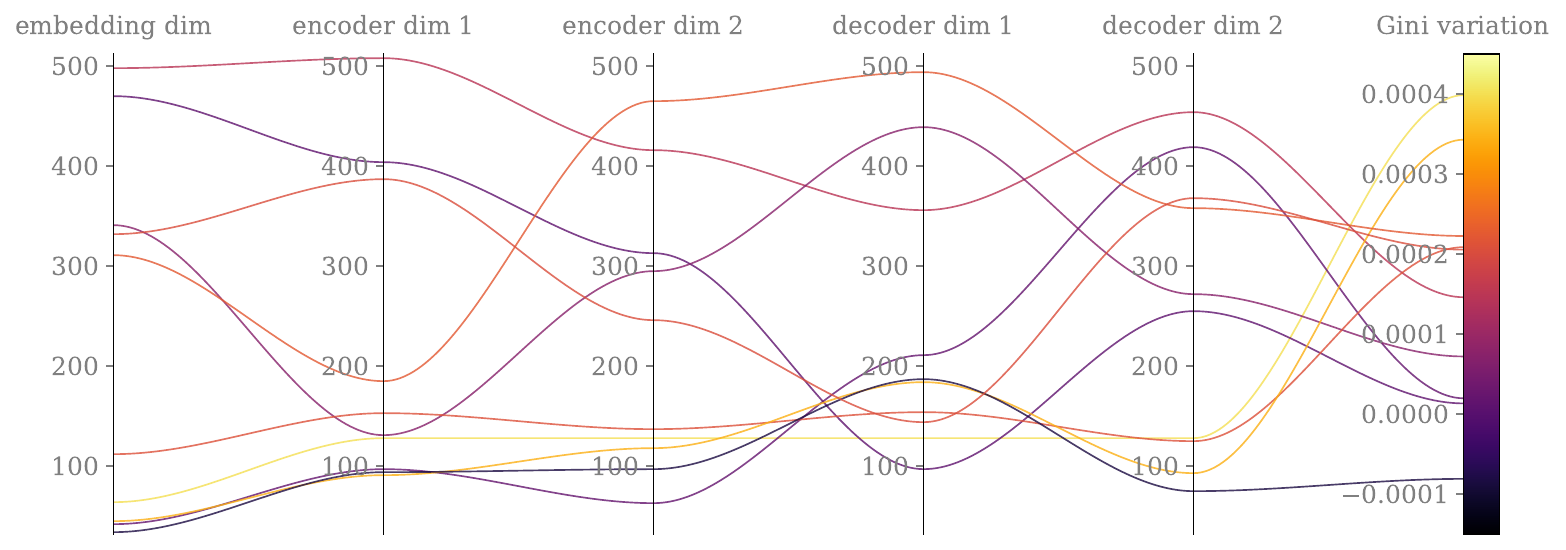}
\caption{TVAE, hard 10\%: parallel coordinates.}
\label{fig:pcp_tvae_hard_amex}
\end{minipage}
\end{figure}

\begin{figure}[!ht]
\begin{minipage}[b]{0.45\textwidth}
\centering
\scalebox{0.85}{
\begin{tabular}{lcccc}
\toprule
Hyperparameter & LB & UB & Default & Best \\ \midrule
Embedding dim. & 32 & 512 & 64 & \textbf{377} \\ 
Discrim. dim. 1 & 32 & 512 & 64 & \textbf{405} \\ 
Discrim. dim. 2 & 32 & 512 & 64 & \textbf{82} \\ 
Generator dim. 1 & 32 & 512 & 64 & \textbf{171} \\ 
Generator dim. 2 & 32 & 512 & 64 & \textbf{118} \\ 
\bottomrule
\end{tabular}
}
\captionof{table}{CTGAN, hard 10\%: tuning setup.}
\label{tab:ctgan_hard_amex}
\end{minipage}
\begin{minipage}[b]{0.55\textwidth}
\centering
\includegraphics[width=.9\textwidth]{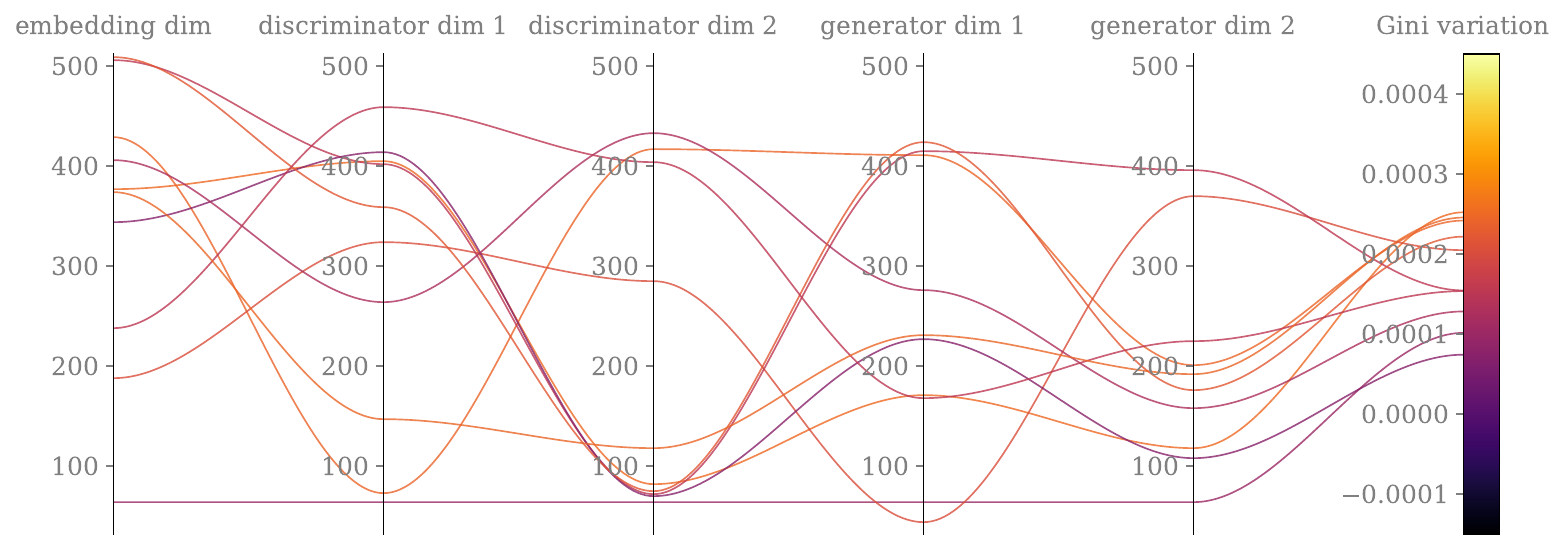}
\caption{CTGAN, hard 10\%: parallel coordinates.}
\label{fig:pcp_ctgan_hard_amex}
\end{minipage}
\end{figure}

\begin{figure}[!ht]
\begin{minipage}[b]{0.45\textwidth}
\centering
\scalebox{0.85}{
\begin{tabular}{lcccc}
\toprule
Hyperparameter & LB & UB & Default & Best \\ \midrule
Embedding dim. & 32 & 512 & 64 & \textbf{479} \\
Encoder dim. 1 & 32 & 512 & 128 & \textbf{249} \\ 
Encoder dim. 2 & 32 & 512 & 128 & \textbf{477} \\ 
Decoder dim. 1 & 32 & 512 & 128 & \textbf{425} \\ 
Decoder dim. 2 & 32 & 512 & 128 & \textbf{33} \\ \bottomrule
\end{tabular}
}
\captionof{table}{TVAE, full data: tuning setup.}
\label{tab:tvae_tot_amex}
\end{minipage}
\begin{minipage}[b]{0.55\textwidth}
\centering
\includegraphics[width=.9\textwidth]{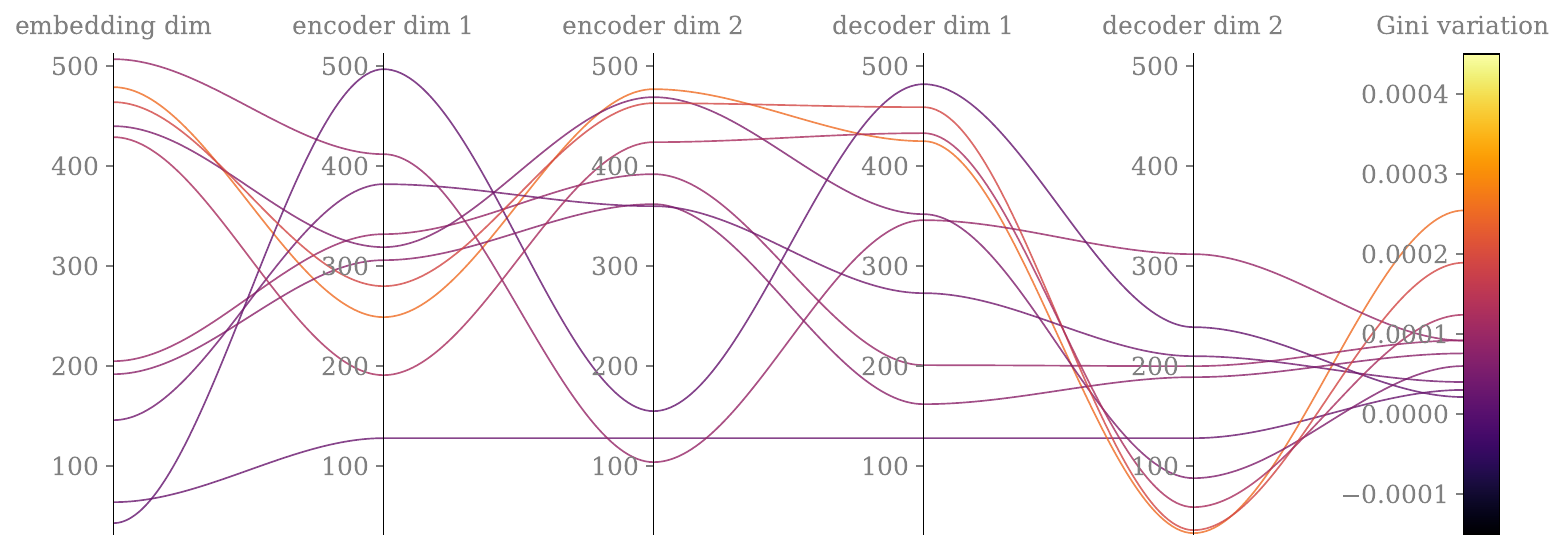}
\caption{TVAE, full data: parallel coordinates.}
\label{fig:pcp_tvae_tot_amex}
\end{minipage}
\end{figure}

\begin{figure}[!ht]
\begin{minipage}[b]{0.45\textwidth}
\centering
\scalebox{0.85}{
\begin{tabular}{lcccc}
\toprule
Hyperparameter & LB & UB & Default & Best \\ \midrule
Embedding dim. & 32 & 512 & 64 & \textbf{64} \\ 
Discrim. dim. 1 & 32 & 512 & 64 & \textbf{64} \\ 
Discrim. dim. 2 & 32 & 512 & 64 & \textbf{64} \\ 
Generator dim. 1 & 32 & 512 & 64 & \textbf{64} \\ 
Generator dim. 2 & 32 & 512 & 64 & \textbf{64} \\
\bottomrule
\end{tabular}
}
\captionof{table}{CTGAN, full data: tuning setup.}
\label{tab:ctgan_tot_amex}
\end{minipage}
\begin{minipage}[b]{0.55\textwidth}
\centering
\includegraphics[width=.9\textwidth]{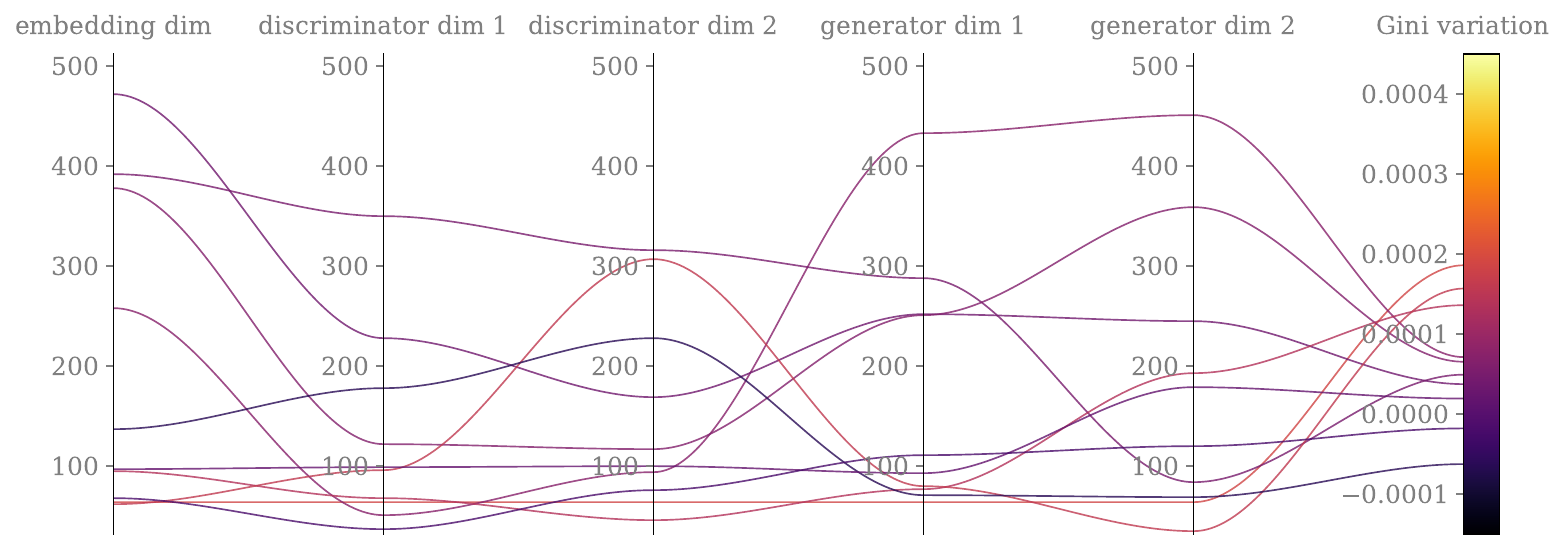}
\caption{CTGAN, full data: parallel coordinates.}
\label{fig:pcp_ctgan_tot_amex}
\end{minipage}
\end{figure}

For TVAE trained on the hardest $10\%$, tuning is performed on the embedding dimension and the architectures of encoder and decoder. We fix the number of hidden layers to two for the encoder and the decoder, tuning the number of units in each hidden layer. Details on the grid search and the best attempt can be found in Table \ref{tab:tvae_hard_amex}, while a parallel coordinates plot is displayed in Figure \ref{fig:pcp_tvae_hard_amex}.
For CTGAN trained on the hardest $10\%$, tuning is performed on the embedding dimension and the discriminator and generator architectures. We fix the number of hidden layers to two for the discriminator and the generator, tuning the number of units in each hidden layer. Further details can be found in Table \ref{tab:ctgan_hard_amex} and Figure \ref{fig:pcp_ctgan_hard_amex}.
Similar summaries are reported for TVAE (Table \ref{tab:tvae_tot_amex}, Figure \ref{fig:pcp_tvae_tot_amex}) and CTGAN (Table \ref{tab:ctgan_tot_amex}, Figure \ref{fig:pcp_ctgan_tot_amex}) on the entire dataset. Notice once again the deterioration in performance with respect to the targeted case.

\subsection{Hyperparameters - Simulation study}\label{sec:a2}

\begin{figure}[!ht]
\begin{minipage}[b]{0.45\textwidth}
\centering
\scalebox{0.85}{
\begin{tabular}{lcccc}
\toprule
Hyperparameter & LB & UB & Default & Best \\ \midrule
Embedding dim. & 16 & 256 & 64 & \textbf{78} \\ 
Encoder dim. 1 & 16 & 256 & 128 & \textbf{210} \\ 
Encoder dim. 2 & 16 & 256 & 128 & \textbf{186} \\ 
Decoder dim. 1 & 16 & 256 & 128 & \textbf{107} \\
Decoder dim. 2 & 16 & 256 & 128 & \textbf{48} \\
\bottomrule
\end{tabular}
}
\captionof{table}{TVAE, hard 5\%: tuning setup.}
\label{tab:tvae_hard_sim}
\end{minipage}
\begin{minipage}[b]{0.55\textwidth}
\centering
\includegraphics[width=.9\textwidth]{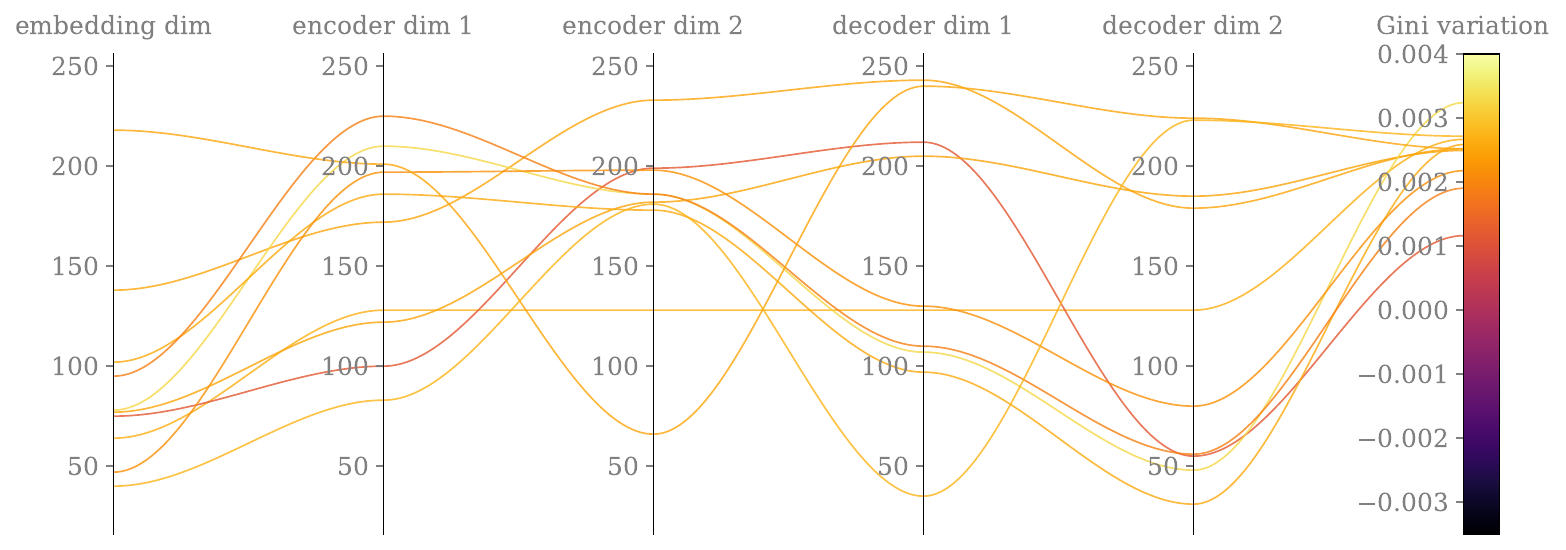}
\caption{TVAE, hard 5\%: parallel coordinates.}
\label{fig:pcp_tvae_hard_sim}
\end{minipage}
\end{figure}

\begin{figure}[!ht]
\begin{minipage}[b]{0.45\textwidth}
\centering
\scalebox{0.85}{
\begin{tabular}{lcccc}
\toprule
Hyperparameter & LB & UB & Default & Best \\ \midrule
Embedding dim. & 16 & 256 & 64 & \textbf{117} \\ 
Encoder dim. 1 & 16 & 256 & 128 & \textbf{104} \\ 
Encoder dim. 2 & 16 & 256 & 128 & \textbf{73} \\ 
Decoder dim. 1 & 16 & 256 & 128 & \textbf{236} \\ 
Decoder dim. 2 & 16 & 256 & 128 & \textbf{250} \\ 
\bottomrule
\end{tabular}
}
\captionof{table}{TVAE, full data: tuning setup.}
\label{tab:tvae_tot_sim}
\end{minipage}
\begin{minipage}[b]{0.55\textwidth}
\centering
\includegraphics[width=.9\textwidth]{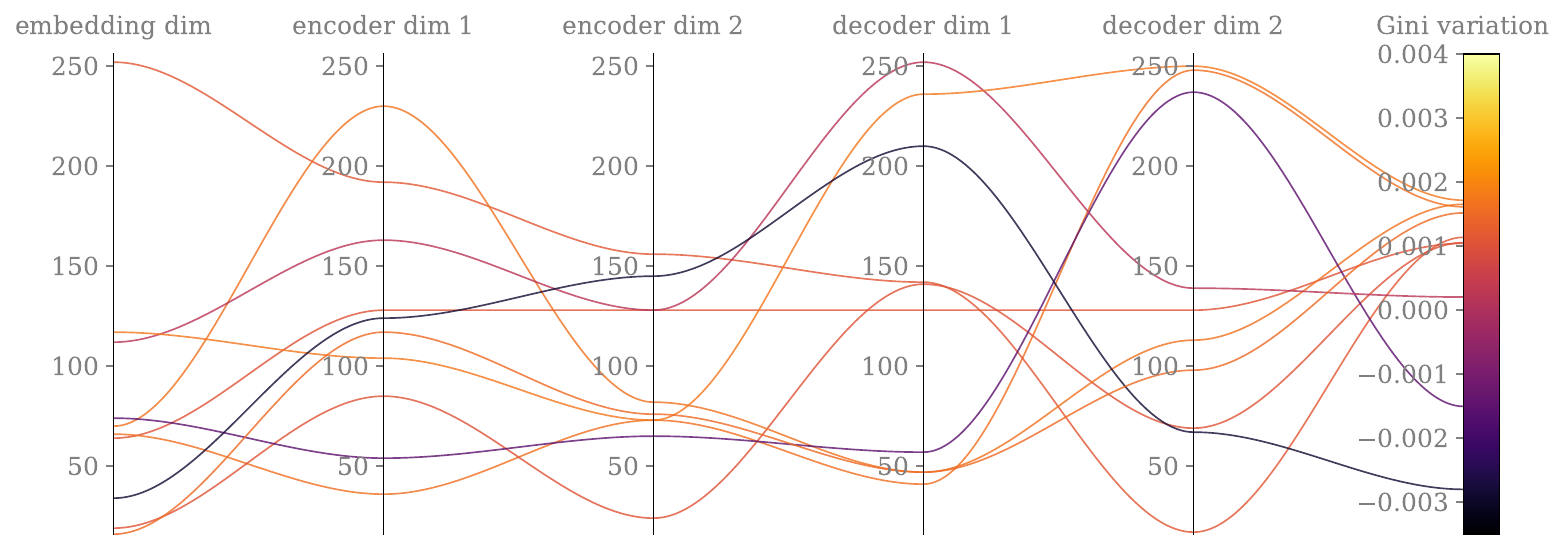}
\caption{TVAE, full data: parallel coordinates.}
\label{fig:pcp_tvae_tot_sim}
\end{minipage}
\end{figure}

The setup is analogous to Appendix \ref{sec:a1}, with the difference that CTGAN has been discarded due to excessive instability during training, and smaller architectures are considered to account for a much simpler dataset. Specifically, Table \ref{tab:tvae_hard_sim} and Figure \ref{fig:pcp_tvae_hard_sim} report the results for TVAE trained on the hardest 5\%, while Table \ref{tab:tvae_tot_sim} and Figure \ref{fig:pcp_tvae_tot_sim} present TVAE trained on the entire dataset. Once again, we notice worse performance in the non-targeted case in the parallel coordinates plots.

\section{Additional plots and results on the Amex dataset}

\subsection{Hardness characterization: a comparison with Data-IQ}

Data-IQ \citep{dataiq} partitions the training data into ``easy'' and ``hard'' points via the concepts of \textit{epistemic} and \textit{aleatoric} uncertainty, combined with a model confidence score calculated from $\mathcal P(\bm{x},\bm{\theta})$, 
representing the probability of assigning a correct label to an observation, given input features $\bm{x}$ and model parameters $\bm{\theta}$. 
The epistemic uncertainty is caused by the variability in the estimated model parameters, whereas the aleatoric uncertainty captures the inherent data uncertainty. Given parameter estimates $\{\bm{\theta}_e\}_{e=1}^E$ obtained during training, a Data-IQ \textit{confidence score} is obtained as $\bar{\mathcal{P}}(\bm{x}) = E^{-1}\sum_{e=1}^E \mathcal{P}(\bm{x}, \bm{\theta}_e)$, whereas the aleatoric uncertainty is given by $v_{\mathrm{al}}(\bm{x}) = E^{-1}\sum_{e=1}^E \mathcal{P}(\bm{x}, \bm{\theta}_e)[1 - \mathcal{P}(\bm{x}, \bm{\theta}_e)]$. Training data can then be partitioned into \textit{(i) easy} points,
corresponding to high $\bar{\mathcal{P}}(\bm{x})$ and low $v_{\mathrm{al}}(\bm{x})$, \textit{(ii) hard} points,
corresponding to low $\bar{\mathcal{P}}(\bm{x})$ and low $v_{\mathrm{al}}(\bm{x})$, and \textit{(iii)} \textit{ambiguous}
points, with high $v_{\mathrm{al}}(\bm{x})$.

\begin{figure}[t]
\centering
\begin{minipage}{0.425\textwidth}
\begin{subfigure}{\textwidth}
\caption{Hardness characterization results}
\label{tab:dataiq}
\centering
\scalebox{0.8}{
\begin{tabular}{cccc}
        \toprule
        Data-IQ tag & Easy & Hard & Ambiguous \\
        \midrule
        $N$ & $284\,902$ & $10\,499$ & $63\,512$ \\
        \% defaulters & $19.07$ & $58.73$ & $51.09$ \\
        \bottomrule
    \end{tabular}
}
\vspace*{.5em}
\end{subfigure}
\begin{subfigure}{\textwidth}
    \centering
    \caption{Joint distribution of $v_{\mathrm{al}}(\bm{x})$ and $\bar{\mathcal{P}}(\bm{x})$}
    \label{fig:hist2d}
    \includegraphics[width=\textwidth]{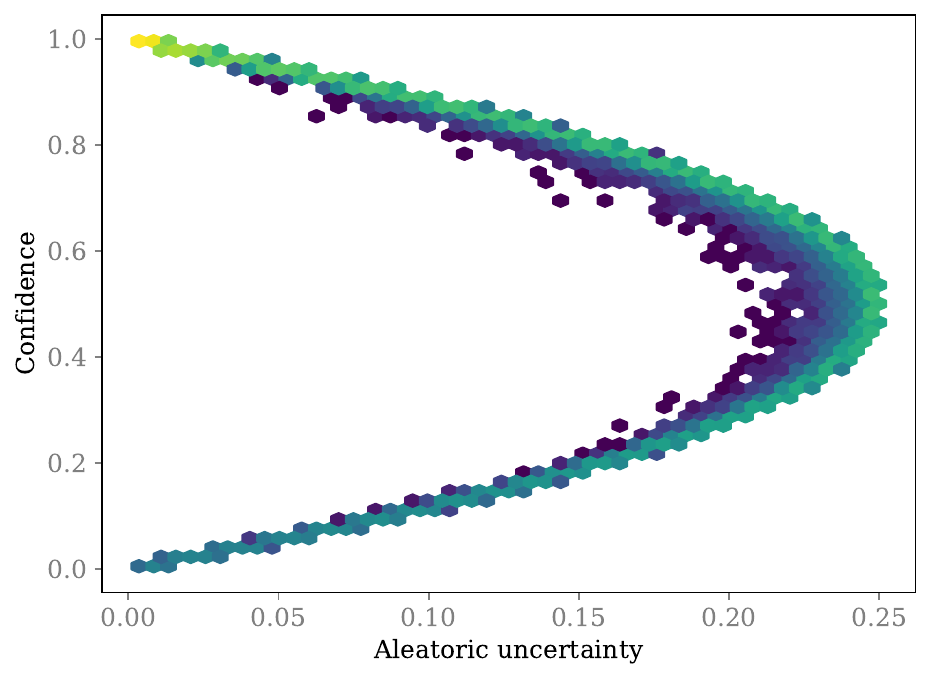}
\end{subfigure}
\end{minipage}
\hspace*{0.025\textwidth}
\begin{minipage}{0.525\textwidth}
\begin{subfigure}{\textwidth}
    \centering
    \caption{Marginal distributions for $v_{\mathrm{al}}(\bm{x})$ and $\bar{\mathcal{P}}(\bm{x})$}
    \label{fig:dataiq_hist}
    \includegraphics[width=\textwidth]{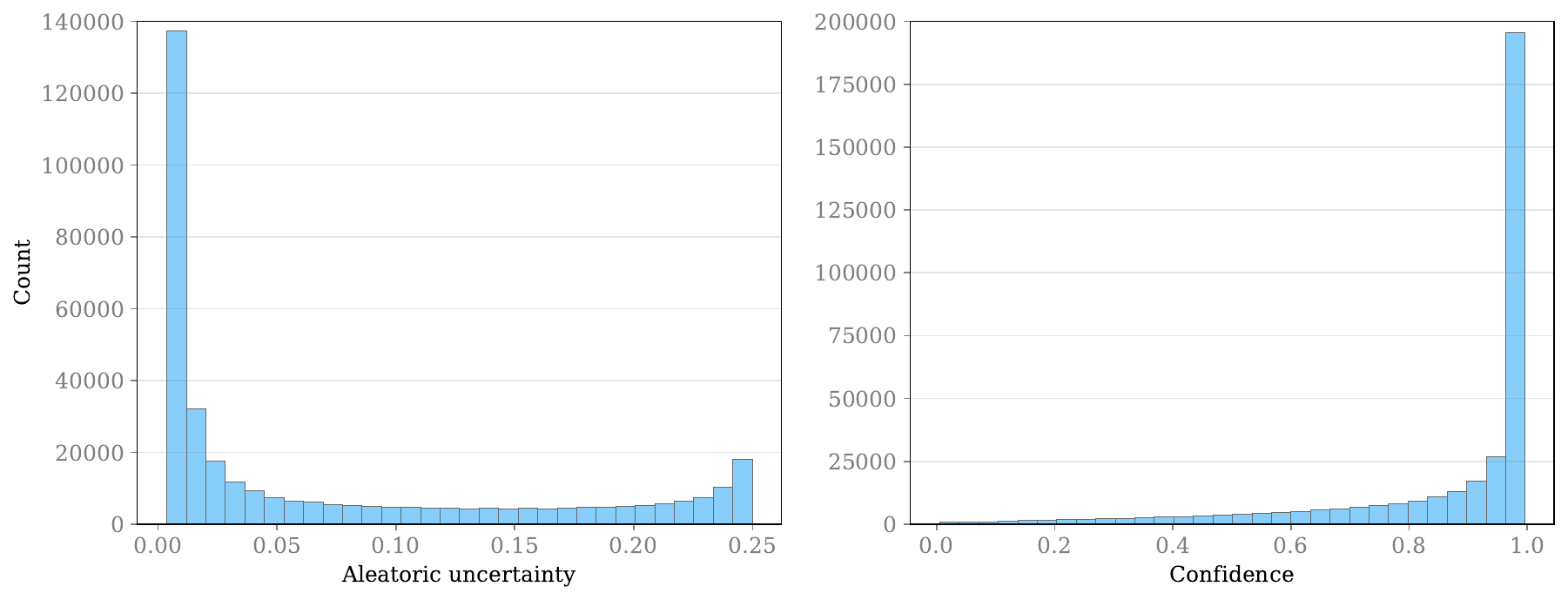}
\end{subfigure}
\begin{subfigure}{\textwidth}
    \centering
    \caption{Feature 0 distribution for defaulters and non-defaulters}
    \label{fig:feat0_dataiq}
    \includegraphics[width=\textwidth]{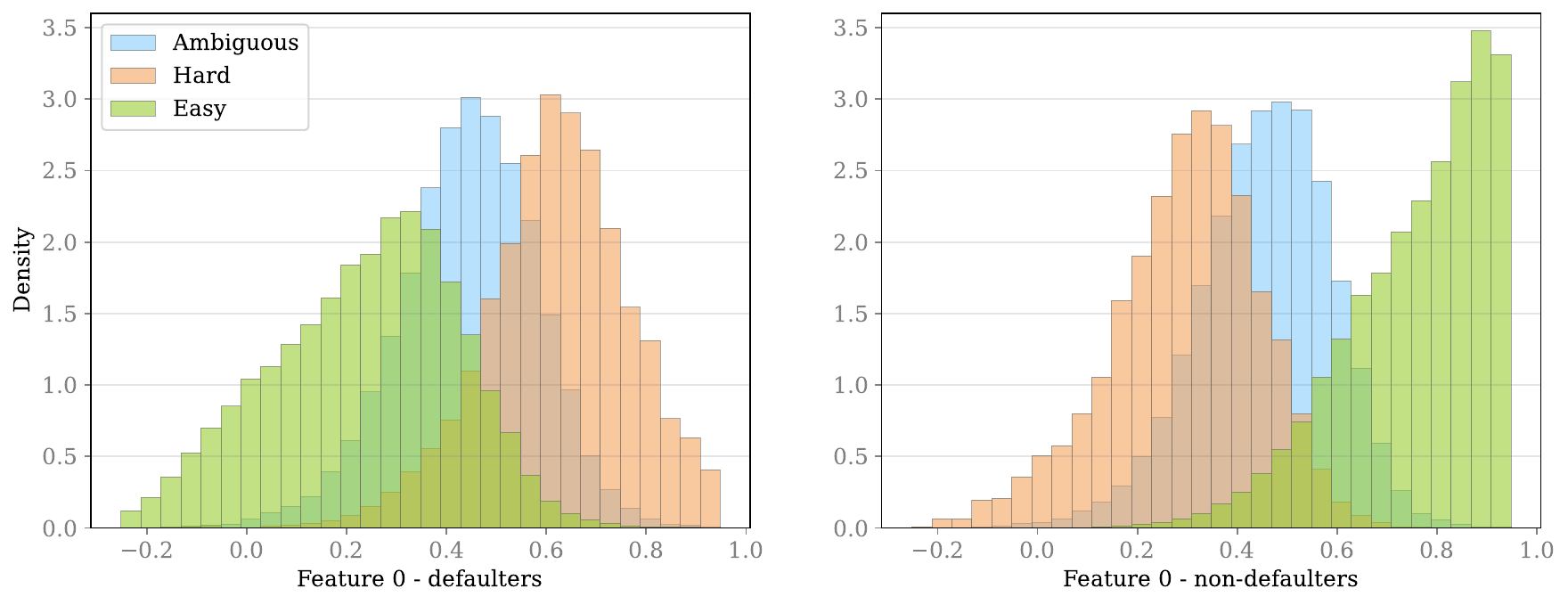}
\end{subfigure}
\end{minipage}
\caption{Data-IQ hardness characterization scores on the Amex credit default prediction dataset.}
\label{fig:dataiq_amex}
\end{figure}

Data-IQ's confidence scores and aleatoric uncertainties were estimated for each training observation in the Amex dataset. Figure \ref{fig:hist2d} displays the relationship between the two: in the top left a large amount of ``easy" points, with high confidence and low aleatoric uncertainty; in the bottom left a few ``hard" points, with low confidence and low aleatoric uncertainty; lastly, the ``ambiguous" points are located around the elbow, with high aleatoric uncertainty. Looking at the marginal distributions in Figure~\ref{fig:dataiq_hist}, it is noticed that XGBoost is generally very confident in its predictions, causing the number of hard points to be low. Setting a low confidence threshold of $0.25$, a high confidence threshold of $0.75$ and a low aleatoric uncertainty threshold of $0.2$ allows to assign to each point a tag in $\{\textit{Easy}, \textit{Hard}, \textit{Ambiguous}\}$. Hard points are less than $3\%$ of training data and both hard and ambiguous points have a much higher proportion of defaulters than the full data (\textit{cf.} Table \ref{tab:dataiq}).

Beyond these initial statistics, it is interesting to know where these points lie in feature space. By looking at the total reduction in loss due to a feature in the nodes where said feature is used to split the data, we can calculate the feature importances of a trained XGBoost classifier. In the case of the Amex dataset, the first feature dominates these scores, with an importance comparable to the bottom 100 features combined. Figure \ref{fig:feat0_dataiq} displays the distribution of this feature separated by Data-IQ tag for both defaulters and non-defaulters: hard defaulters somewhat overlap with easy non-defaulters and vice-versa, while ambiguous defaulters are almost indistinguishable from ambiguous non-defaulters. This is a common issue with tabular data, referred to as \textit{outcome heterogeneity} and is known to be captured by Data-IQ \citep{dataiq}.

\subsection{Synthetic data generation}

\begin{figure}[b]
    \centering
    \begin{subfigure}{0.475\textwidth}
        \centering
        \caption{CTGAN}
        \label{fig:ctgan_hard_hist}
        \includegraphics[width=\textwidth]{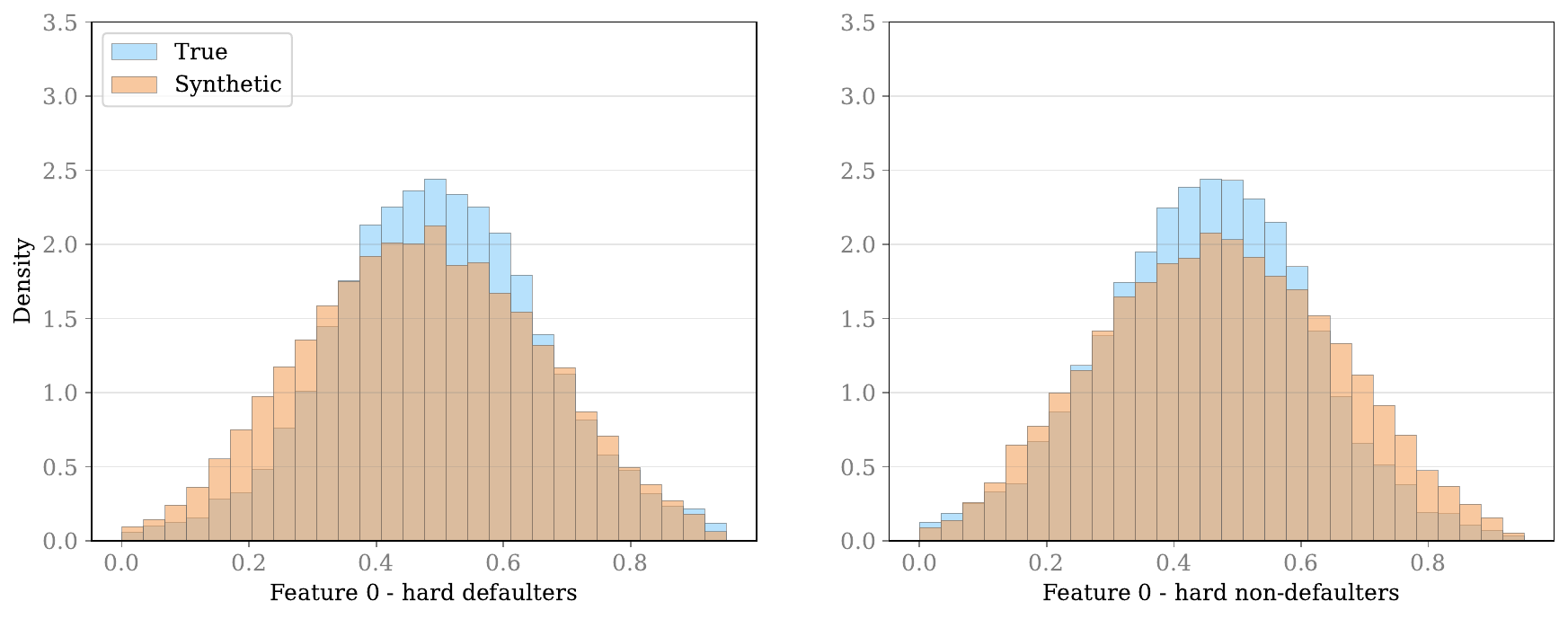}
    \end{subfigure}
    \hspace{.025\textwidth}
    \begin{subfigure}{0.475\textwidth}
        \centering
        \caption{TVAE}
        \label{fig:tvae_hard_hist}
        \includegraphics[width=\textwidth]{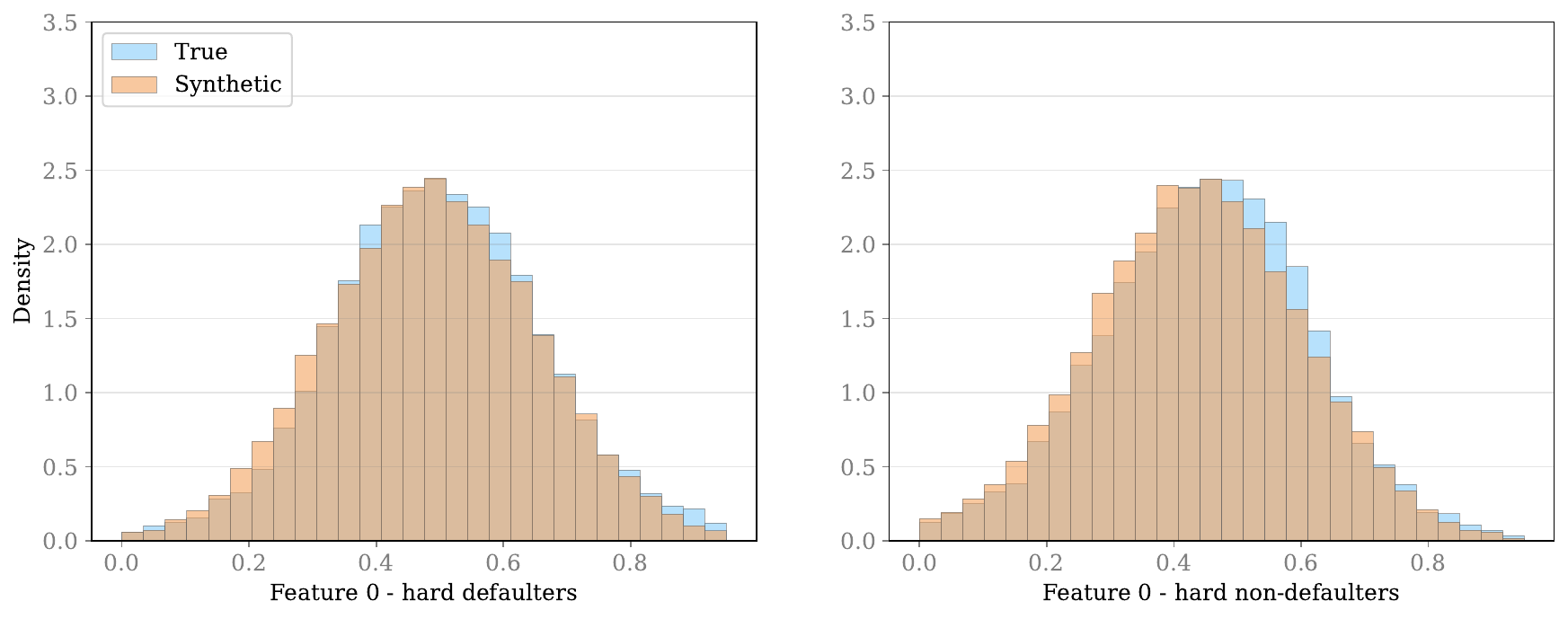}
    \end{subfigure}
    \caption{Histogram of feature 0 in the Amex dataset for true and synthetic data generated via CTGAN and TVAE, restricted to the 10\% harest data points according to the 100NN Shapley scores.}
    \label{fig:hard_hist}
\end{figure}

Both CTGAN and TVAE are implemented in Python using a heavily customized version of the \texttt{sdv} package, with extensive details about the training given in Appendix~\ref{sec:ap_training}.
The quality of the synthetic samples can be examined directly for the first feature: Figure \ref{fig:ctgan_hard_hist} (CTGAN) displays a good but not perfect overlap with real data, whereas from Figure \ref{fig:tvae_hard_hist} (TVAE) we can see better overlap of synthetic and real data with respect to CTGAN for the first feature. 

